\documentclass[sigconf]{acmart}
\AtBeginDocument{%
  }

\setcopyright{acmlicensed}
\copyrightyear{2018}
\acmYear{2018}
\acmDOI{XXXXXXX.XXXXXXX}
\acmConference[Conference acronym 'XX]{Make sure to enter the correct
  conference title from your rights confirmation email}{June 03--05,
  2018}{Woodstock, NY}
\acmISBN{978-1-4503-XXXX-X/2018/06}

\usepackage{multirow}
\usepackage{tikz}
\usepackage{amsmath}
\usepackage{amsthm}
\usepackage{bbding}
\usepackage[normalem]{ulem}
\useunder{\uline}{\ul}{}
\usepackage{adjustbox}

\begin{document}

\title{MMAFFBen: A Multilingual and Multimodal Affective Analysis Benchmark for Evaluating LLMs and VLMs }

\author{Zhiwei Liu}
\affiliation{%
  \institution{The University of Manchester}
  \country{United Kingdom}
}
\email{zhiwei.liu@manchester.ac.uk}

\author{Lingfei Qian}
\affiliation{%
  \institution{The Fin AI}
    \country{Singapore}
}
\email{lfqian94@gmail.com}

\author{Qianqian Xie}
\affiliation{%
  \institution{School of Artificial Intelligence, Wuhan University}
    \country{China}
}
\email{xqq.sincere@gmail.com}

\author{Jimin Huang}
\affiliation{%
  \institution{The Fin AI}
    \country{Singapore}
}
\email{jimin.huang@thefin.ai}

\author{Kailai Yang}
\affiliation{%
  \institution{The University of Manchester}
  \country{United Kingdom}
}
\email{kailai.yang@manchester.ac.uk}

\author{Sophia Ananiadou}
\affiliation{%
  \institution{The University of Manchester}
  \country{United Kingdom}
}
\email{sophia.ananiadou@manchester.ac.uk}

\renewcommand{\shortauthors}{Zhiwei Liu et al.}

\begin{abstract}
Large language models and vision-language models (which we jointly call LMs) have transformed NLP and CV, demonstrating remarkable potential across various fields. However, their capabilities in affective analysis (i.e. sentiment analysis and emotion detection) remain underexplored. This gap is largely due to the absence of comprehensive evaluation benchmarks, and the inherent complexity of affective analysis tasks. In this paper, we introduce MMAFFBen, the first extensive open-source benchmark for multilingual multimodal affective analysis. MMAFFBen encompasses text, image, and video modalities across 35 languages, covering four key affective analysis tasks: sentiment polarity, sentiment intensity, emotion classification, and emotion intensity. Moreover, we construct the MMAFFIn dataset for fine-tuning LMs on affective analysis tasks, and further develop MMAFFLM-3b and MMAFFLM-7b based on it. We evaluate various representative LMs, including GPT-4o-mini, providing a systematic comparison of their affective understanding capabilities. This project is publicly released to foster transparent, reproducible, and inclusive progress in affective analysis studies and applications\footnote{https://github.com/lzw108/MMAFFBen\\https://huggingface.co/datasets/lzw1008/MMAFFBen/tree/main\\https://huggingface.co/datasets/lzw1008/MMAFFIn/tree/main}.
\end{abstract}




\maketitle

\section{Introduction}

Recently, large language models (LLMs) such as ChatGPT and GPT-4 \cite{achiam2023gpt}, as well as vision-language models (VLMs) like Qwen-2.5-VL \cite{bai2025qwen2}, GPT-4o \cite{hurst2024gpt}, and InternVL \cite{chen2024internvl}, have demonstrated outstanding performance in various fields. In this paper, we collectively refer to LLMs and VLMs as large models (LMs) that can follow instructions. These models excel not only in natural language understanding and generation tasks \cite{karanikolas2023large} but also in tasks involving image-text comprehension \cite{shuai2024survey}, multimodal question answering \cite{pramanick2024spiqa}, and scene text understanding tasks \cite{liao2024vlm2scene}. Recent studies \cite{liu2024emollms,yang2024emollm,yang2025omni} have shown that LMs demonstrate great potential in affective analysis (i.e. sentiment analysis and emotion detection). However, a comprehensive understanding of their capabilities and limitations in this field is still lacking. This is mainly due to the shortage of evaluation research and benchmarks, as well as the complexity and specialized nature of affective analysis tasks.

With the diverse development of social media, people who speak various languages read or post text, images, or video content on social media platforms \cite{aguero2021deep,chandrasekaran2021multimodal}. Therefore, it is urgently necessary to develop a benchmark that comprehensively evaluates the affective analysis capabilities of LMs. Many studies have actively contributed to this domain. Liu et al. \cite{liu2024emollms} develop an affective evaluation benchmark (AEB) with 14 text-based tasks from various sources to evaluate LLMs, including sentiment polarity (SP), sentiment intensity (SI), emotion classification (EC) and emotion intensity (EI) tasks. Emily et al. \cite{ohman2020xed} introduce XED, a multilingual dataset for fine-grained EC, comprising human-annotated Finnish sentences and English sentences, along with projected annotations in 30 additional languages. MMS \cite{augustyniak2023massively} is a multilingual SP task, covering 27 languages representing 6 language families. Li et al. \cite{li2024facial} propose FABA-bench, a benchmark for facial affective behavior analysis that evaluates both recognition and generation capabilities of VLMs. Kosti et al. \cite{kosti2020context} present EMOTIC, a dataset of real-world images annotated with apparent emotions and the continuous dimensions of valence, arousal, and dominance. FindingEmo \cite{mertens2024findingemo} contains 25k images, focusing on complex, multi-person scenes in naturalistic social settings. Each image is annotated across valence, arousal, and discrete emotion labels. MERR \cite{cheng2024emotion} is a video-based dataset comprising 28,618 coarse-grained and 4,487 fine-grained annotated samples across diverse emotional categories. Bujnowski et al. \cite{bujnowski2024samsemo} present SAMSEMO, a dataset for multimodal and multilingual emotion recognition based on 23k raw video scenes in five languages (EN, DE, ES, PL, KO), collected from diverse domains. CMU-MOSEI \cite{zadeh2018multimodal} consists of 23,453 annotated video segments from 1,000 distinct speakers across 250 topics. Each segment includes manually aligned transcriptions down to the phoneme level, along with SI and EI labels. Additional dataset contributions are summarized in Table \ref{tab:relatedbenchmark}.

However, existing affective analysis evaluation benchmarks, such as AEB \cite{liu2024emollms}, MMS \cite{augustyniak2023massively},  FABA \cite{li2024facial}, MERR \cite{cheng2024emotion} as shown in Table \ref{tab:relatedbenchmark}, cover only a limited range of evaluation tasks and primarily focus on a single task (either sentiment analysis or emotion detection). Moreover, they are mostly restricted to a single modality (e.g., text, image, or video) or a single language (mainly English). Some datasets (e.g. SAMSEMO \cite{bujnowski2024samsemo}, CMU-MOSEI \cite{zadeh2018multimodal}) convert audio to text based on raw video data, which is referred to as multimodal. However, the final modality they actually address is still video. While in the real world, modalities are a mixture of various types of raw data. These limitations hinder their ability to comprehensively evaluate LMs on diverse and fine-grained affective analysis tasks (e.g. sentiment intensity and emotion intensity). Consequently, these benchmarks are insufficient to fully reveal the potential capabilities and limitations of LMs in affective analysis domain.

To bridge the existing evaluation gaps, we propose MMAFFBen – a comprehensive open-source benchmark for multilingual multimodal affective analysis. Developed through the collaborative efforts of experts in computer science and linguistics, MMAFFBen encompasses text, image, and video modalities across 35 languages, and supports evaluation for four affective analysis tasks: Emotion Intensity (EI), Emotion Classification (EC), Sentiment Polarity (SP), and Sentiment Intensity (SI). Each category targets specific aspects of affective data processing and analysis, ensuring a thorough evaluation of LLMs and VLMs and demonstrating their proficiency in handling complex affective scenarios. Moreover, we also build an instruction-tuning dataset MMAFFIn for fine-tuning LMs. Applying MMAFFIn, we develop MMAFFLM-3B and MMAFFLM-7B based on Qwen2.5-VL-3B-Instruct and Qwen2.5-VL-7B-Instruct. Based on MMAFFBen, we access 20 representative LMs, including general LLMs, VLMs, affective analysis LLMs, and MMAFFLM-3b and MMAFFLM-7b.

In summary, the main contributions of this paper are: 1) we propose MMAFFBen, the first comprehensive multilingual and multimodal open-sourced evaluation benchmark for LMs in the affective analysis domain, 2) we propose MMAFFIn, the first instruction-tuning dataset for comprehensive multilingual and multimodal affective analysis. 3) Based on MMAFFIn, we develop two MMAFF models for affective analysis, and 4) we conduct systematic evaluation of 20 LMs using MMAFFBen, showcasing their limitations and advantages and highlighting directions for future work.

\begin{table}[!t]
\footnotesize
\caption{Comparison of different affective analysis benchmarks. ``EI'': emotion intensity, ``EC'': emotion classification, ``SP'': sentiment polarity, ``SI'': sentiment intensity.}
  \label{tab:relatedbenchmark}
\resizebox{0.48\textwidth}{!}{
\begin{tabular}{lccccccc}
\hline
{\color[HTML]{212121} }      & EI                                  & EC                                  & SP                                  & SI                                  & Multi-Lingual                       & Raw Modality       & Support LMs?                        \\ \hline
AEB \cite{liu2024emollms}                         & \textcolor{green}{\large \checkmark} & \textcolor{green}{\large \checkmark} & \textcolor{green}{\large \checkmark} & \textcolor{green}{\large \checkmark} & \textcolor{red}{\small \XSolidBrush}                        & Text               & \textcolor{green}{\large \checkmark} \\
XED \cite{ohman2020xed}                         &  \textcolor{red}{\small \XSolidBrush}                       & \textcolor{green}{\large \checkmark} & \textcolor{red}{\small \XSolidBrush}                        & \textcolor{red}{\small \XSolidBrush}                        & \textcolor{green}{\large \checkmark} & Text               & \textcolor{red}{\small \XSolidBrush}                        \\
MMS \cite{augustyniak2023massively}                         & \textcolor{red}{\small \XSolidBrush}                        & \textcolor{red}{\small \XSolidBrush}                        & \textcolor{green}{\large \checkmark} & \textcolor{red}{\small \XSolidBrush}                        & \textcolor{green}{\large \checkmark} & Text               & \textcolor{red}{\small \XSolidBrush}                        \\ \hline
FABA \cite{li2024facial}     & \textcolor{red}{\small \XSolidBrush}                        & \textcolor{green}{\large \checkmark} & \textcolor{red}{\small \XSolidBrush}                        & \textcolor{red}{\small \XSolidBrush}                        & \textcolor{red}{\small \XSolidBrush}                        & Image              & \textcolor{green}{\large \checkmark} \\
EMOTIC \cite{kosti2020context}                       & \textcolor{red}{\small \XSolidBrush}                        & \textcolor{green}{\large \checkmark} & \textcolor{red}{\small \XSolidBrush}                        & \textcolor{green}{\large \checkmark} & \textcolor{red}{\small \XSolidBrush}                        & Image              & \textcolor{red}{\small \XSolidBrush}                        \\
HECO \cite{yang2022emotion}                         & \textcolor{red}{\small \XSolidBrush}                        & \textcolor{green}{\large \checkmark} & \textcolor{red}{\small \XSolidBrush}                        & \textcolor{green}{\large \checkmark} & \textcolor{red}{\small \XSolidBrush}                        & Image              & \textcolor{red}{\small \XSolidBrush}                        \\
FindingEmo \cite{mertens2024findingemo}                  & \textcolor{red}{\small \XSolidBrush}                        & \textcolor{green}{\large \checkmark} & \textcolor{red}{\small \XSolidBrush}                        & \textcolor{green}{\large \checkmark} & \textcolor{red}{\small \XSolidBrush}                        & Image              & \textcolor{red}{\small \XSolidBrush}                        \\
FER2013 \cite{FER2013}                     & \textcolor{red}{\small \XSolidBrush}                        & \textcolor{green}{\large \checkmark} & \textcolor{red}{\small \XSolidBrush}                        & \textcolor{red}{\small \XSolidBrush}                        & \textcolor{red}{\small \XSolidBrush}                        & Image              & \textcolor{red}{\small \XSolidBrush}                        \\
CFAPS \cite{gong2011revision}                       & \textcolor{green}{\large \checkmark} & \textcolor{green}{\large \checkmark} & \textcolor{red}{\small \XSolidBrush}                        & \textcolor{red}{\small \XSolidBrush}                        & \textcolor{red}{\small \XSolidBrush}                        & Image              & \textcolor{red}{\small \XSolidBrush}                        \\ \hline
MERR \cite{cheng2024emotion} & \textcolor{red}{\small \XSolidBrush}                        & \textcolor{green}{\large \checkmark} & \textcolor{red}{\small \XSolidBrush}                        & \textcolor{red}{\small \XSolidBrush}                        & \textcolor{red}{\small \XSolidBrush}                        & Video              & \textcolor{green}{\large \checkmark} \\
SAMSEMO \cite{bujnowski2024samsemo}                      & \textcolor{red}{\small \XSolidBrush}                        & \textcolor{green}{\large \checkmark} & \textcolor{red}{\small \XSolidBrush}                        & \textcolor{red}{\small \XSolidBrush}                        & \textcolor{green}{\large \checkmark} & Video              & \textcolor{red}{\small \XSolidBrush}                        \\
CMU-MOSEI \cite{zadeh2018multimodal}                   & \textcolor{green}{\large \checkmark} & \textcolor{red}{\small \XSolidBrush}                        & \textcolor{red}{\small \XSolidBrush}                        & \textcolor{green}{\large \checkmark} & \textcolor{red}{\small \XSolidBrush}                        & Video              & \textcolor{red}{\small \XSolidBrush}                        \\
MELD  \cite{poria2019meld}                        & \textcolor{red}{\small \XSolidBrush}                        & \textcolor{green}{\large \checkmark} & \textcolor{green}{\large \checkmark} & \textcolor{red}{\small \XSolidBrush}                        & \textcolor{red}{\small \XSolidBrush}                        & Video              & \textcolor{red}{\small \XSolidBrush}                        \\
IEMOCAP \cite{busso2008iemocap}                      & \textcolor{red}{\small \XSolidBrush}                        & \textcolor{green}{\large \checkmark} & \textcolor{red}{\small \XSolidBrush}                        & \textcolor{green}{\large \checkmark} & \textcolor{red}{\small \XSolidBrush}                        & Video              & \textcolor{red}{\small \XSolidBrush}                        \\
CHSIMS  \cite{yu2020ch}                     & \textcolor{red}{\small \XSolidBrush}                        & \textcolor{red}{\small \XSolidBrush}                        & \textcolor{green}{\large \checkmark} & \textcolor{green}{\large \checkmark} & \textcolor{red}{\small \XSolidBrush}                        & Video              & \textcolor{red}{\small \XSolidBrush}                        \\ \hline
MMAFFBen                     & \textcolor{green}{\large \checkmark} & \textcolor{green}{\large \checkmark} & \textcolor{green}{\large \checkmark} & \textcolor{green}{\large \checkmark} & \textcolor{green}{\large \checkmark} & Text, Image, Video & \textcolor{green}{\large \checkmark} \\ \hline
\end{tabular}
}
\end{table}

\section{MMAFFBen}


With the rapid rise of social media, people around the world interact not only through text but also via images and videos \cite{chandrasekaran2021multimodal}. A thorough and systematic approach is essential for evaluating the capabilities of LMs. We propose a comprehensive benchmark for affective evaluation tasks,
categorizing and assessing LLMs and VLMs across various modalities and languages. Multimodality is a trending topic. We select data from three modalities—text, image, and video—which are also the most popular forms of communication on social media. With the widespread adoption of the internet, people worldwide can communicate through social media without being limited to a specific language \cite{aguero2021deep}. Therefore, we selected data from 34 languages for text, data from both Western and Eastern countries for images, and data in 6 languages for videos (A total of 35 distinct
languages). In this section, we explore the details of MMAFFBen, including the details of data sources, tasks, construction process, and evaluation metrics. 

\subsection{Data Sources}

MMAFFBen's evaluation tasks are constructed using open-source datasets from prior research, which were initially designed for evaluation scenarios that did not involve LMs. To adapt these datasets for LLM evaluation, we carefully curated a diverse set of prompts and restructured the data into instruction-response pairs. This transformation ensures that the datasets are suitable for assessing LMs in a zero-shot setting. By leveraging these reformulated datasets, MMAFFBen provides a comprehensive framework for evaluating the capabilities of LMs across various tasks.

\subsection{Tasks}

\begin{table*}[]
\footnotesize
\caption{The tasks, datasets, data statistics, and evaluation metrics included in MMAFFBen. ``multi'' denotes multi-class task. ``single'' denotes single label task. For SP (number), ``number'' denotes the number of the categories. A total of 35 distinct languages are included across all datasets.}
  \label{tab:datastatisctic}
\begin{tabular}{clllp{4cm}ll}
\hline
\multicolumn{1}{l}{{\color[HTML]{212121} Modality}} & Dataset                               & Task                             & Language           & test                                          & Evaluation        & License         \\ \hline
                                                    & SemEva2018   \cite{SemEval2018Task1}  & EI                & en, ar, es         & en: 4068, ar: 1563, es: 2616                  & pcc               & Limit           \\
                                                    & SemEva2018   \cite{SemEval2018Task1}  & EC (multi)   & en, ar, es         & en: 3259, ar: 1518, es: 2854                  & jac, mi-F1, ma-F1 & Limit           \\
                                                    & SemEva2018   \cite{SemEval2018Task1}  & SP (7)           & en, ar, es         & en: 937, ar: 730, es: 648                     & pcc               & Limit           \\
                                                    & SemEva2018   \cite{SemEval2018Task1}  & SI              & en, ar, es         & en: 937, ar: 730, es: 648                     & pcc               & Limit           \\
                                                    & EWECT-usual \cite{SMP2020-EWECT}                           & EC (single)  & zh                 & 5000                                          & acc, ma-F1        & public          \\
                                                    & EWECT-virus \cite{SMP2020-EWECT}                           & EC (single)  & zh                 & 3000                                          & acc, ma-F1        & public          \\
                                                    & Onlineshopping \cite{online_shopping_10_cats}                       & SP (2)           & zh                 & 2500                                          & acc, ma-F1        & public          \\
                                                    & XED   \cite{ohman2020xed}             & EC (multi)   & 25                 & 500 for each language                        & jac, mi-F1, ma-F1 & CC-BY           \\
\multirow{-10}{*}{Text}                             & MMS   \cite{augustyniak2023massively} & SP (3)           & 27                 & 500 for each language                        & acc, ma-F1        & CC-BY-NC 4.0    \\ \hline
                                                    & EMOTIC \cite{kosti2020context}        & EC (multi)   & -                  & 7203                                          & jac, mi-F1, ma-F1 & MIT             \\
                                                    & EMOTIC   \cite{kosti2020context}      & SI              & -                  & 7203                                          & pcc               & MIT             \\
                                                    & FER2013 \cite{FER2013}                              & EC (single)  & -                  & 7178                                          & acc, ma-F1        & DbCL v1.0       \\
                                                    & CFAPS   \cite{gong2011revision}       & EC (single)  & -                  & 174                                           & acc, ma-F1        & public          \\
\multirow{-5}{*}{image}                             & CFAPS   \cite{gong2011revision}       & EI                & -                  & 174                                           & pcc               & public          \\ \hline
                                                    & SANSEMO   \cite{bujnowski2024samsemo} & EC (multi)   & de, en, es, ko, ps & de: 357, en: 585, es: 276, ko:   186, pl: 769 & jac, mi-F1, ma-F1 & CC BY-NC-SA 4.0 \\
                                                    & CHSIMS   \cite{yu2020ch}             & SP (3)           & zh                 & 457                                           & acc, ma-F1        & MIT             \\
                                                    & CHSIMS   \cite{yu2020ch}             & SI              & zh                 & 457                                           & pcc               & MIT             \\
                                                    & MELD   \cite{poria2019meld}           & EC (single)  & en                 & 2530                                          & acc, ma-F1        & GPL-3.0 license \\
\multirow{-5}{*}{video}                             & MELD   \cite{poria2019meld}           & SP (3)           & en                 & 2530                                          & acc, ma-F1        & GPL-3.0 license \\ \hline
\end{tabular}
\end{table*}

Table \ref{tab:datastatisctic} presents all tasks, datasets, data statistics, and evaluation metrics covered by MMAFFBen.

Affective analysis is typically divided into sentiment analysis and emotion detection. Sentiment analysis contains sentiment polarity classification and sentiment intensity tasks \cite{tian2018polarity}. Emotion detection includes emotion classification and emotion intensity tasks \cite{dixon2015emotion}.

\textbf{Sentiment polarity (SP)}: Polarity reflects the overall tone of the content, indicating its positivity, negativity, or neutrality \cite{devika2016sentiment}. In MMAFFBen, there are three kinds of polarity tasks, binary classification (i.e. negative and positive in Onlineshopping \cite{online_shopping_10_cats}), three classifications (i.e. negative, neutral, and positive in MMS \cite{augustyniak2023massively}, CHSIMS \cite{yu2020ch}, MELD \cite{poria2019meld}), and 7 fine-grained classifications (i.e. from very negative to very positive in SemEval2018 \cite{SemEval2018Task1}).

\textbf{Sentiment intensity (SI)}: Sentiment intensity is represented by a real value to indicate the strength of polarity, allowing for better capture of dynamic sentiment changes \cite{hutto2014vader}.  MMAFFBen, CHSIMS \cite{yu2020ch}, EMOTIC \cite{kosti2020context}, and SemEval2018 \cite{SemEval2018Task1} include sentiment intensity tasks. SemEval2018 focuses on the valence dimension, while EMOTIC extends beyond valence by also providing intensity measurements for arousal and dominance. For consistency, we normalize all sentiment intensity values to the range of -1 to 1.

\textbf{Emotion classification (EC)}: Emotion detection aims to identify specific emotion categories in the content \cite{gaind2019emotion}. Unlike sentiment analysis, emotion detection typically involves finer-grained emotion categories, such as happiness, anger, etc \cite{acheampong2020text}. Categorical models define a fixed set of discrete emotional states. Examples include Shaver's \cite{shaver1987emotion} (sadness, love, joy, anger, surprise, and fear), Ekman \cite{ekman1992argument}  (joy, anger, fear, disgust, sadness, and surprise), and Plutchik \cite{plutchik1980general} (anticipation, surprise, anger, fear, trust, disgust, joy, and sadness). In MMAFFBen, there are two kinds of emotion classifications, including single-class classification (i.e. only one emotion category is assigned to the content) in EWECT \cite{SMP2020-EWECT}, FER2013 \cite{FER2013}, CFAPS \cite{gong2011revision}, and MELD \cite{poria2019meld} and multi-class classification (i.e. one or more emotions are assigned to the content) in SemEval2018 \cite{SemEval2018Task1}, XED \cite{ohman2020xed}, EMOTIC \cite{kosti2020context} and SANSEMO \cite{bujnowski2024samsemo}.

\textbf{Emotion intensity (EI)}: Emotion intensity expresses the strength of a particular emotion through a real value, enabling a more precise and nuanced assessment of emotional expressions \cite{reisenzein2024measuring}. In MMAFFBen, SemEval2018 \cite{SemEval2018Task1} and CFAPS \cite{gong2011revision} include the emotion intensity. For consistency, we normalize the emotion intensity values to the range [0,1].

\subsection{Dataset Details and Process}

\textbf{(1) Text}
 
\textbf{SemEval2018 \cite{SemEval2018Task1}}\footnote{If you use this dataset, please refer to: https://competitions.codalab.org/competitions\\/17751\#learn\_the\_details-terms\_and\_conditions.}: The dataset is based on SemEval-2018 Task 1: Affect in Tweets, which includes a series of highly annotated affective analysis subtasks. It contains emotion intensity, emotion classification, sentiment polarity, sentiment intensity, and emotion intensity classification. We select the first four tasks in MMAFFBen. For EI, it annotates the intensity of anger, fear, joy, and sadness with a real-value score ranging from 0 (least) to 1 (most). For EC, it has 11 emotions (i.e. anger, anticipation, disgust, fear, joy, love, optimism, pessimism, sadness, surprise, trust) and neutral. For SP and SI, it annotates the valence dimension. In SP, the data has 7 fine-grained polity labels (very negative, moderately negative, slightly negative, neutral, slightly positive, moderately positive, very positive). In SI, it also uses one real-value score between 0 (most negative) and 1 (positive) to annotate the content. For consistency, we normalize the score to [-1,1]. The dataset includes English, Arabic, and Spanish.

\textbf{EWECT \cite{SMP2020-EWECT}:} EWECT is collected from the Evaluation of Weibo Emotion Classification Technology. It consists of two parts: the first contains randomly collected posts from Chinese Weibo platform covering various general topics (EWECT-usual); the second is a pandemic Weibo dataset, where all posts are related to COVID-19 (EWECT-virus). In the dataset, each item is categorized into one of six emotion categories: happiness, sadness, anger, fear, surprise, and neutral. 

\textbf{Onlineshopping \cite{online_shopping_10_cats}: } Onlineshopping is a Chinese dataset collected from e-commerce platforms, covering multiple domains such as books, tablets, mobile phones, fruits, shampoo, water heaters, Mengniu, clothing, computers, and hotels. It is categorized into two classes: negative and positive.

\textbf{XED \cite{ohman2020xed}:} XED dataset consists of emotion-annotated movie subtitles from OPUS. It is a multi-class dataset, using Plutchik’s eight core emotions for annotation. The original annotations are primarily from English and Finnish, while the remaining annotations were created by projecting them onto aligned subtitles in other languages. In MMAFFBen, we selected 25 languages with more than 2,000 entries.

\textbf{MMS \cite{augustyniak2023massively}:} MMS dataset is selected from over 350 datasets reported in scientific literature based on strict quality standards. This corpus covers 27 languages from 6 language families. It includes three labels: negative, neutral, and positive.

\textbf{(2) Image}

\textbf{EMOTIC \cite{kosti2020context}:} EMOTIC is collected from COCO \cite{lin2014microsoft}, Ade20k \cite{zhou2019semantic}, and the Internet, containing images of people in various natural situations, with their apparent emotions. The EMOTIC dataset combines two different types of affective representations: (1) a set of 26 discrete categories and (2) dimensions of valence, arousal, and dominance. To make the model more focused on affective tasks, we cropped the images using the provided bounding boxes. 

\textbf{FER2013 \cite{FER2013}:} The FER2013 dataset consists of 48x48 pixel grayscale images of human faces. The faces are centered and occupy approximately the same amount of space. The data is categorized into one of six emotions (anger, disgust, fear, happiness, sadness, surprise) based on facial expressions or neutral.

\textbf{CFAPS \cite{gong2011revision}:} CFAPS is a Chinese facial emotion image dataset annotated by 100 university students, covering six emotion types (happiness, sadness, anger, fear, surprise, disgust) and neutral. The dataset also includes emotion intensity ratings ranging from 1 to 9 (1 being the weakest and 9 the strongest). For consistency, we normalized the intensity scores to a range of 0 to 1.

\textbf{(3) Video}

\textbf{SANSEMO \cite{bujnowski2024samsemo}:} SANSEMO consists of 23k video clips in five languages (EN, DE, ES, PL
and KO), featuring approximately 1,400 speakers across various video types. All video clips have been manually transcribed and annotated with emotions. It is a multi-label task, including the following emotion categories: happiness, sadness, anger, fear, surprise, disgust, neutral, or others.

\textbf{CHSIMS \cite{yu2020ch}:} CHSIMS is a Chinese multi-modal dataset collected from movies,
TV series, and variety shows. It contains 2,281 refined video segments. For each clip, the sentiment state is labeled as negative, neutral, or positive. One of {-1.0, -0.8, -0.6, -0.4, -0.2, 0.0, 0.2, 0.4, 0.6, 0.8, 1.0} is used to indicate sentiment intensity.

\textbf{MELD \cite{poria2019meld}:} MELD is a dialogue dataset based on EmotionLines \cite{hsu2018emotionlines}, featuring over 800 dialogues and more than 9,000 utterances from the TV show Friends. Each conversation involves multiple speakers. Every utterance is labeled with one of seven emotions: happiness, sadness, anger, fear, surprise, disgust, or neutral. In addition, MELD includes sentiment annotations (positive, negative, or neutral) for each utterance.

For all the aforementioned datasets, if the original dataset is already split, we adopt the default test set for constructing MAFFBen. If the dataset is not pre-split, 20\% of the data is randomly sampled to serve as the evaluation set for MAFFBen. We construct instructions adapted for LMs based on the following template. \textit{[Task description]} is the task-specific prompt for the different datasets. \textit{Raw content} is the raw text or transcription. More details (all prompts for each dataset) can be found in Table \ref{tab:promptdescriptions}.

\begin{center}
\footnotesize
\fcolorbox{black}{gray!5}{
\begin{minipage}{0.45\textwidth}
\footnotesize
\textbf{Template for constructing instruction}  \\
\textbf{Task:} \textit{[Task description]}  \\
\textbf{Content: } \textit{Raw content}
\end{minipage}
}
\end{center}

\subsection{Metrics}

For SemEval, we follow the official evaluation metrics, using the Pearson correlation coefficient (pcc) for EI, SP, and SI. For the remaining datasets, pcc is also used for SI and EI. For SP and EC,  we report the macro F1 (ma-F1).

\section{MMAFFIn and MMAFFLM}

After constructing the MMAFFBen, we apply the remaining data of each dataset to build MMAFFIn to support LMs' instruction tuning. Table \ref{tab:trainstatistic} shows the statistics of MMAFFIn. Due to computational resource constraints, we extracted at most 2,000 samples from each dataset in MMAFFIn to train the models (using all available samples for datasets with fewer than 2,000 entries). We built MMAFFLM-3b and MMAFFLM-7b by fine-tuning Qwen2.5-VL-3B-Instruct and Qwen2.5-VL-7B-Instruct based on LLaMA-Factory framework \cite{zheng2024llamafactory}. We trained the models for one epoch, using a streaming strategy with a batch size of 256 and a learning rate of 5e-6. All models are trained on two Nvidia A100 GPUs, each with 80GB of memory.

\begin{table}[]
\footnotesize
\caption{Statistics of MMAFFIn.}
  \label{tab:trainstatistic}
\begin{tabular}{lll}
\hline
Modality               & Dataset           & Train/Val                                   \\ \hline
\multirow{9}{*}{Text}  & SemEva2018-EI     & en: 7102/1464. ar: 3376/661. es:   4541/793 \\
                       & SemEva2018-EC     & en: 6838/886. ar: 2278/585. es:   3559/679  \\
                       & SemEva2018-SP     & en: 1181/449. ar: 932/138. es:   1566/229   \\
                       & SemEva2018-SI     & en: 1181/449. ar: 932/138. es:   1566/229   \\
                       & EWECT-usual   -EC & 10000/2000                                  \\
                       & EWECT-virus-EC    & 8606/2000                                   \\
                       & Onlineshopping-SP & 8000/1500                                   \\
                       & XED-EC            & 25000/12500                                 \\
                       & MMS-SP            & 27000/13500                                 \\ \hline
\multirow{5}{*}{image} & EMOTIC-EC         & 23266/3315                                  \\
                       & EMOTIC-SI         & 23266/3315                                  \\
                       & FER2013-EC        & 22967/5742                                  \\
                       & CFAPS-EC          & 556/140                                     \\
                       & CFAPS-EI          & 556/140                                     \\ \hline
\multirow{5}{*}{video} & SANSEMO-EC        & 10165/2178                                  \\
                       & CHSIMS-SP         & 1368/456                                    \\
                       & CHSIMS-SI         & 1368/456                                    \\
                       & MELD-EC           & 9699/1073                                   \\
                       & MELD-SP           & 9699/1073                                   \\ \hline
Total                  &                   & 216568/56088                                \\ \hline
\end{tabular}
\end{table}

\section{Evaluation}

\begin{table*}[]
\caption{Results on SemEval-2018 datasets. Except for the Ec task, all others use the pcc metric. ``ave'' denotes the average pcc of anger, fear, joy and sadness in EI task in SemEval. ``jac'' denotes Jaccard score. ``mi'' denotes micro. ``ma'' denotes macro. ``Overall'' denotes the average of pcc/ma-F1 of each task. ``LM-T'' denotes models only supporting text. ``LM-TI'' denotes models only supporting text and image. ``LM-TV'' denotes models only supporting text and video. ``LM-TIV'' denotes models supporting text, image, and video.}
  \label{tab:resultsSemEval}
\resizebox{\textwidth}{!}{
\begin{tabular}{clccccccccccccccccccc}
\hline
\multicolumn{1}{l}{}     & \multirow{3}{*}{Models} & \multicolumn{6}{c}{SemEval\_en}                                                               & \multicolumn{6}{c}{SemEval\_ar}                                                               & \multicolumn{6}{c}{SemEval\_es}                                                               & \multirow{3}{*}{Overall} \\
\multicolumn{1}{l}{}     &                         & EI            & SP            & SI            & \multicolumn{3}{c}{Ec}                        & EI            & SP            & SI            & \multicolumn{3}{c}{Ec}                        & EI            & SP            & SI            & \multicolumn{3}{c}{Ec}                        &                          \\
\multicolumn{1}{l}{}     &                         & ave           & valence       & valence       & jac           & mi-F1         & ma-F1         & ave           & valence       & valence       & jac           & mi-F1         & ma-F1         & ave           & valence       & valence       & jac           & mi-F1         & ma-F1         &                          \\ \hline
\multirow{4}{*}{LM-T}    & EmoLlama-chat-7b        & \textbf{79.9} & \textbf{87.3} & \textbf{85.6} & \textbf{60.9} & \textbf{72.4} & \textbf{59.2} & 43.0          & 66.1          & 74.5          & 36.2          & 52.6          & 38.3          & \textbf{73.7} & {\ul 74.4}    & \textbf{81.7} & {\ul 40.1}    & {\ul 56.3}    & \textbf{45.7} & 67.4                     \\
                         & Llama3.2-1b-instruct    & 14.4          & 27.1          & 34.8          & 36.8          & 51.0          & 39.2          & 6.3           & 18.7          & 22.0          & 15.0          & 25.5          & 20.9          & 13.0          & 22.5          & 38.1          & 22.6          & 34.6          & 27.6          & 23.7                     \\
                         & Llama3.2-3b-instruct    & 54.7          & 65.2          & 71.1          & 32.7          & 45.8          & 35.8          & 26.6          & 58.5          & 67.7          & 33.5          & 47.6          & 37.2          & 43.8          & 53.7          & 64.4          & 29.1          & 42.7          & 35.1          & 51.1                     \\
                         & Mistral-7b-instruct     & 58.4          & 77.7          & 70.8          & 30.7          & 43.0          & 35.6          & 35.9          & 70.9          & 64.1          & 30.6          & 42.1          & 33.9          & 57.6          & 73.4          & 56.7          & 29.8          & 40.7          & 34.2          & 55.8                     \\ \hline
\multirow{3}{*}{LM-TI}   & Llama3.2-11b-instruct            & 58.1          & 69.6          & 67.4          & 33.5          & 47.5          & 37.7          & 46.7          & 72.4          & 72.8          & 31.9          & 46.7          & 34.9          & 48.3          & 66.1          & 65.3          & 28.8          & 41.6          & 33.7          & 56.1                     \\
                         & Llava-1.5-7b            & 34.0          & 48.6          & 56.3          & 24.2          & 35.8          & 30.7          & 18.3          & 45.0          & 47.0          & 17.3          & 27.8          & 22.7          & 30.8          & 45.8          & 53.2          & 21.8          & 32.6          & 28.0          & 38.4                     \\
                         & Llava-1.5-13b           & 52.5          & 72.3          & 63.7          & 26.0          & 38.1          & 29.9          & 28.1          & 65.8          & 59.3          & 21.2          & 31.6          & 24.6          & 48.4          & 66.7          & 53.7          & 27.1          & 39.3          & 31.7          & 49.7                     \\ \hline
\multirow{2}{*}{LM-TV}   & Llava-nextvideo-7b      & 12.2          & 32.1          & 54.2          & 18.1          & 30.0          & 23.8          & 0.6           & 29.5          & 25.3          & 11.7          & 20.6          & 15.4          & 6.4           & 30.5          & 40.9          & 18.9          & 31.3          & 24.1          & 24.6                     \\
                         & Llava-nextvideo-7b-dpo  & 8.6           & 35.3          & 61.9          & 18.9          & 31.2          & 24.6          & -2.4          & 24.8          & 35.3          & 11.2          & 20.3          & 14.4          & 2.9           & 29.6          & 42.1          & 20.1          & 32.9          & 23.8          & 25.1                     \\ \hline
\multirow{11}{*}{LM-TIV} & Qwen2.5-vl-3b           & 54.5          & 71.2          & 38.3          & 28.2          & 42.0          & 30.1          & 42.5          & 72.0          & 31.2          & 25.2          & 37.3          & 28.6          & 57.5          & 64.8          & 42.5          & 23.9          & 37.6          & 26.6          & 46.7                     \\
                         & Qwen2.5-vl-7b           & 36.8          & 69.2          & 36.8          & 22.1          & 35.0          & 26.0          & 21.6          & 61.4          & 26.4          & 22.2          & 34.8          & 24.9          & 29.0          & 58.3          & 24.4          & 23.8          & 37.1          & 28.3          & 36.9                     \\
                         & InternVL2.5-1b          & 13.9          & 55.1          & 38.1          & 25.7          & 33.6          & 29.1          & 8.3           & 46.7          & 32.3          & 24.8          & 33.7          & 29.1          & 11.2          & 46.9          & 33.0          & 19.1          & 26.9          & 23.4          & 30.6                     \\
                         & InternVL2.5-2b          & 34.4          & 52.0          & 17.0          & 25.5          & 38.2          & 28.9          & 9.7           & 35.0          & 26.9          & 21.0          & 32.6          & 25.7          & 16.4          & 43.5          & 8.7           & 19.1          & 31.7          & 24.6          & 26.9                     \\
                         & InternVL2.5-8b          & 58.1          & 77.3          & 39.9          & 34.2          & 46.4          & 35.7          & 34.1          & 65.1          & 25.3          & 28.5          & 38.9          & 32.3          & 56.3          & 56.1          & 19.8          & 28.8          & 40.3          & 32.5          & 44.4                     \\
                         & InternVL2.5-1b-MPO      & 9.6           & 46.6          & 15.4          & 30.5          & 37.5          & 32.5          & 3.8           & 39.9          & 26.3          & 28.3          & 36.8          & 32.1          & 4.4           & 42.8          & 12.6          & 22.9          & 30.2          & 26.1          & 24.3                     \\
                         & InternVL2.5-2b-MPO      & 39.9          & 46.3          & 12.4          & 25.8          & 39.0          & 28.5          & 8.8           & 42.1          & 10.8          & 21.0          & 32.0          & 23.3          & 24.0          & 28.6          & 9.7           & 22.1          & 36.3          & 26.1          & 25.0                     \\
                         & InternVL2.5-8b-MPO      & 59.5          & 62.8          & 27.9          & 36.4          & 48.4          & 38.2          & 37.1          & 50.6          & 12.9          & 30.6          & 42.6          & 34.5          & 56.7          & 40.8          & 13.2          & 31.1          & 43.1          & 34.9          & 39.1                     \\
                         & GPT-4o-mini             & {\ul 71.4}    & {\ul 82.4}    & 82.0          & 42.1          & 58.6          & 44.6          & \textbf{63.9} & \textbf{85.4} & \textbf{86.4} & \textbf{46.5} & \textbf{64.1} & \textbf{47.1} & {\ul 73.0}    & \textbf{81.4} & {\ul 80.6}    & \textbf{42.1} & \textbf{60.0} & {\ul 44.7}    & \textbf{70.2}            \\
                         & MMAFFLM-3b              & 63.6          & 76.8          & 82.5          & {\ul 46.6}    & {\ul 61.6}    & {\ul 46.9}    & 57.7          & 78.2          & {\ul 82.4}    & {\ul 44.8}    & {\ul 59.9}    & {\ul 45.1}    & 63.5          & 73.6          & 79.1          & 38.4          & 53.2          & 43.1          & 66.1                     \\
                         & MMAFFLM-7b              & 70.3          & 77.8          & {\ul 83.2}    & 43.2          & 59.3          & 42.3          & {\ul 62.1}    & {\ul 78.4}    & 81.8          & 39.7          & 57.4          & 40.5          & 69.8          & 73.0          & 79.6          & 35.7          & 52.3          & 38.4          & {\ul 66.4}               \\ \hline
\end{tabular}
}
\end{table*}

\begin{table*}[]
\footnotesize
\caption{Results on other text datasets, image datasets, and video datasets. For SI and EI tasks, we present the pcc. For SP and EC, we show the ma-F1. ``LM-T'' denotes models only supporting text. ``LM-TI'' denotes models only supporting text and image. ``LM-TV'' denotes models only supporting text and video. ``LM-TIV'' denotes models supporting text, image, and video.}
  \label{tab:resultsall}
\resizebox{\textwidth}{!}{
\begin{tabular}{@{}clccccccccccccccccccccc@{}}
\toprule
\multicolumn{1}{l}{}     &                                             & \multicolumn{6}{c}{Text}                                                                                            & \multicolumn{8}{c}{Image}                                                                                                                          & \multicolumn{6}{c}{Video}                                                                                          & All           \\
\multicolumn{1}{l}{}     & \multicolumn{1}{l|}{Models}                 & \adjustbox{angle=90}{EWECT-usual}   & \adjustbox{angle=90}{EWECT-virus}   & \adjustbox{angle=90}{Onlineshopping} & \adjustbox{angle=90}{MMS}           & \adjustbox{angle=90}{XED}           & \multicolumn{1}{c|}{\adjustbox{angle=90}{Average}}       & \adjustbox{angle=90}{EMOTIC-EC}     & \adjustbox{angle=90}{EMOTIC-SI-V}   & \adjustbox{angle=90}{EMOTIC-SI-A}   & \adjustbox{angle=90}{EMOTIC-SI-D}   & \adjustbox{angle=90}{FER2013}       & \adjustbox{angle=90}{CFAPS-EC}      & \adjustbox{angle=90}{CFAPS-EI}      & \multicolumn{1}{c|}{\adjustbox{angle=90}{Average}}       & \adjustbox{angle=90}{SAMSEMO}       & \adjustbox{angle=90}{MELD-EC}       & \adjustbox{angle=90}{MELD-SP}       & \adjustbox{angle=90}{CHSIMS-SP}     & \adjustbox{angle=90}{CHSIMS-SI}     & \multicolumn{1}{c|}{\adjustbox{angle=90}{Average}}       & \adjustbox{angle=90}{Average}       \\ \midrule
\multirow{4}{*}{LM-T}    & \multicolumn{1}{l|}{EmoLlama-chat-7b}       & 45.6          & 30.5          & 44.0           & \textbf{48.6}          & 20.3          & \multicolumn{1}{c|}{37.8}          & -             & -             & -             & -             & -             & -             & -             & \multicolumn{1}{c|}{-}             & -             & -             & -             & -             & -             & \multicolumn{1}{c|}{-}             & -             \\
                         & \multicolumn{1}{l|}{Llama3.2-1b-instruct}   & 24.2          & 18.0          & 49.1           & 30.9          & 18.7          & \multicolumn{1}{c|}{28.2}          & -             & -             & -             & -             & -             & -             & -             & \multicolumn{1}{c|}{-}             & -             & -             & -             & -             & -             & \multicolumn{1}{c|}{-}             & -             \\
                         & \multicolumn{1}{l|}{Llama3.2-3b-instruct}   & 51.9          & 39.9          & 52.2           & 32.9          & 9.1           & \multicolumn{1}{c|}{37.2}          & -             & -             & -             & -             & -             & -             & -             & \multicolumn{1}{c|}{-}             & -             & -             & -             & -             & -             & \multicolumn{1}{c|}{-}             & -             \\
                         & \multicolumn{1}{l|}{Mistral-7b-instruct}    & 45.1          & 36.5          & 55.4           & 33.1          & {\ul 23.3}    & \multicolumn{1}{c|}{38.7}          & -             & -             & -             & -             & -             & -             & -             & \multicolumn{1}{c|}{-}             & -             & -             & -             & -             & -             & \multicolumn{1}{c|}{-}             & -             \\ \midrule
\multirow{3}{*}{LM-TI}   & \multicolumn{1}{l|}{Llama3.2-11b-instruct}           & 42.3          & 35.8          & 57.4           & 37.5          & 20.0          & \multicolumn{1}{c|}{38.6}          & 18.7          & 23.7          & 22.2          & 12.9          & 34.1          & 36.6          & 20.4          & \multicolumn{1}{c|}{24.1}          & -             & -             & -             & -             & -             & \multicolumn{1}{c|}{-}             & -             \\
                         & \multicolumn{1}{l|}{Llava-1.5-7b}           & 41.6          & 29.4          & 55.5           & 32.8          & 23.1          & \multicolumn{1}{c|}{36.5}          & 19.2          & 32.5          & 21.7          & 16.4          & 39.6          & 41.6          & 13.9          & \multicolumn{1}{c|}{26.4}          & -             & -             & -             & -             & -             & \multicolumn{1}{c|}{-}             & -             \\
                         & \multicolumn{1}{l|}{Llava-1.5-13b}          & 40.9          & 32.4          & 56.5           & 36.5          & 22.4          & \multicolumn{1}{c|}{37.7}          & 12.5          & 25.0          & 12.9          & 9.7           & 32.8          & 25.7          & 19.9          & \multicolumn{1}{c|}{19.8}          & -             & -             & -             & -             & -             & \multicolumn{1}{c|}{-}             & -             \\ \midrule
\multirow{2}{*}{LM-TV}   & \multicolumn{1}{l|}{Llava-nextvideo-7b}     & 36.7          & 27.9          & 57.1           & 30.7          & 17.9          & \multicolumn{1}{c|}{34.1}          & -             & -             & -             & -             & -             & -             & -             & \multicolumn{1}{c|}{-}             & 15.6          & 17.3          & 30.1          & 36.6          & 25.0          & \multicolumn{1}{c|}{24.9}          & -             \\
                         & \multicolumn{1}{l|}{Llava-nextvideo-7b-dpo} & 37.7          & 29.8          & 57.1           & 31.7          & 19.0          & \multicolumn{1}{c|}{35.1}          & -             & -             & -             & -             & -             & -             & -             & \multicolumn{1}{c|}{-}             & 15.4          & 18.2          & 31.8          & 38.0          & 48.9          & \multicolumn{1}{c|}{30.5}          & -             \\ \midrule
\multirow{11}{*}{LM-TIV} & \multicolumn{1}{l|}{Qwen2.5-vl-3b}          & 41.5          & 37.0          & 57.6           & 35.9          & 13.7          & \multicolumn{1}{c|}{37.1}          & 12.3          & 33.6          & 34.0          & 16.9          & 8.2           & 20.1          & 38.7          & \multicolumn{1}{c|}{23.4}          & 31.0          & 23.2          & 43.0          & 39.3          & 38.0          & \multicolumn{1}{c|}{34.9}          & 30.8          \\
                         & \multicolumn{1}{l|}{Qwen2.5-vl-7b}          & 46.3          & 38.1          & 56.2           & 31.3          & 12.8          & \multicolumn{1}{c|}{36.9}          & 13.8          & 47.0          & {\ul 44.6}    & 23.4          & 32.5          & 36.9          & 8.9           & \multicolumn{1}{c|}{29.6}          & 30.9          & 29.2          & 39.3          & 33.2          & 47.6          & \multicolumn{1}{c|}{36.1}          & 33.7          \\
                         & \multicolumn{1}{l|}{InternVL2.5-1b}         & 22.0          & 22.7          & 57.8           & 33.3          & 18.4          & \multicolumn{1}{c|}{30.8}          & 12.9          & 17.2          & 4.1           & -2.0          & 29.7          & 30.2          & -1.0          & \multicolumn{1}{c|}{13.0}          & 22.7          & 16.4          & 26.4          & 25.5          & 22.1          & \multicolumn{1}{c|}{22.6}          & 21.1          \\
                         & \multicolumn{1}{l|}{InternVL2.5-2b}         & 52.8          & 42.0          & 58.3           & 32.4          & 4.6           & \multicolumn{1}{c|}{38.0}          & 23.0          & 6.5           & 6.5           & 1.6           & 21.1          & 18.9          & 13.7          & \multicolumn{1}{c|}{13.0}          & 27.5          & 13.8          & 34.6          & 39.0          & 44.9          & \multicolumn{1}{c|}{31.9}          & 26.0          \\
                         & \multicolumn{1}{l|}{InternVL2.5-8b}         & 52.5          & 37.8          & 59.6           & 34.8          & 9.7           & \multicolumn{1}{c|}{38.9}          & 23.0          & 39.9          & 32.4          & 20.6          & 42.3          & 40.3          & 21.9          & \multicolumn{1}{c|}{31.5}          & 36.8          & 32.0          & 39.6          & 56.4          & 64.1          & \multicolumn{1}{c|}{45.8}          & 37.9          \\
                         & \multicolumn{1}{l|}{InternVL2.5-1b-MPO}     & 30.8          & 28.7          & 56.9           & 30.9          & 21.1          & \multicolumn{1}{c|}{33.7}          & 17.2          & 17.7          & 10.9          & -2.4          & 36.8          & 31.1          & -10.8         & \multicolumn{1}{c|}{14.3}          & 29.7          & 18.2          & 32.0          & 23.4          & 26.8          & \multicolumn{1}{c|}{26.0}          & 23.5          \\
                         & \multicolumn{1}{l|}{InternVL2.5-2b-MPO}     & 50.9          & 39.1          & 55.0           & 29.4          & 11.1          & \multicolumn{1}{c|}{37.1}          & {\ul 25.1}    & 10.3          & 17.2          & 6.3           & 32.5          & 35.6          & 11.3          & \multicolumn{1}{c|}{19.8}          & 33.6          & 19.3          & 27.5          & 31.2          & 42.3          & \multicolumn{1}{c|}{30.8}          & 28.1          \\
                         & \multicolumn{1}{l|}{InternVL2.5-8b-MPO}     & 51.1          & 35.7          & 56.2           & 31.4          & 12.5          & \multicolumn{1}{c|}{37.4}          & 24.6          & \textbf{55.8} & \textbf{55.8} & {\ul 26.3}    & {\ul 43.2}    & 40.2          & 47.6          & \multicolumn{1}{c|}{{\ul 41.9}}    & 39.3          & 34.2          & 34.8          & 54.2          & {\ul 64.3}    & \multicolumn{1}{c|}{45.4}          & 41.6          \\
                         & \multicolumn{1}{l|}{GPT-4o-mini}            & \textbf{69.5} & 57.6          & 61.9           & \textbf{48.6} & 21.2          & \multicolumn{1}{c|}{51.8}          & \textbf{28.9} & {\ul 52.0}    & \textbf{55.8} & \textbf{31.8} & \textbf{58.3} & \textbf{61.0} & 47.2          & \multicolumn{1}{c|}{\textbf{47.9}} & {\ul 40.4}    & 26.4          & 40.5          & 56.7          & \textbf{70.3} & \multicolumn{1}{c|}{46.9}          & {\ul 48.7}    \\
                         & \multicolumn{1}{l|}{MMAFFLM-3b}             & 66.9          & \textbf{60.3} & \textbf{93.9}  & {\ul 43.3}    & \textbf{26.5} & \multicolumn{1}{c|}{\textbf{58.2}} & 17.9          & 35.0          & 20.2          & 19.5          & 33.7          & 58.3          & {\ul 71.7}    & \multicolumn{1}{c|}{36.6}          & 39.7          & \textbf{36.2} & {\ul 61.4}    & {\ul 58.0}    & 56.2          & \multicolumn{1}{c|}{{\ul 50.3}}    & 47.0          \\
                         & \multicolumn{1}{l|}{MMAFFLM-7b}             & {\ul 67.6}    & {\ul 58.2}    & {\ul 93.7}     & 43.0          & 22.8          & \multicolumn{1}{c|}{{\ul 57.1}}    & 16.7          & 41.8          & 35.9          & 19.0          & 37.5          & {\ul 58.4}    & \textbf{73.0} & \multicolumn{1}{c|}{40.3}          & \textbf{40.8} & {\ul 35.6}    & \textbf{65.7} & \textbf{59.9} & 63.2          & \multicolumn{1}{c|}{\textbf{53.0}} & \textbf{49.0} \\ \bottomrule
\end{tabular}
}
\end{table*}

We evaluate the zero-shot performance of 20 representative general LMs and affective analysis LLMs on the MMAFFBen benchmark, including: 

1) Models only supporting text (LM-T):

Llama3.2-(1b,3b)-instruct \cite{meta2024llama}: The Llama 3.2 instruction-tuned text only models that are optimized for multilingual dialogue use cases, including agentic retrieval and summarization tasks. Mistral-7b-instruct \cite{jiang2023mistral4}: A 7.3 billion parameter language model that represents a major advance in LLMs' capabilities. EmoLlama-chat-7b \cite{liu2024emollms}: One of Emotional LLMs with the best overall performance for comprehensive English text affective analysis.

2) Models only supporting text and image (LM-TI):

Llama3.2-11b-instruct \cite{meta2024llama}: A Llama 3.2-Vision instruction-tuned model that is optimized for visual recognition, image reasoning, captioning, and answering general questions about images. Llava1.5-(7b,13b) \cite{liu2023improvedllava,liu2023llava}: Open-source chatbot models by fine-tuning LLaMA/ Vicuna on GPT-generated multimodal instruction-following data.

3) Models only supporting text and video (LM-TV):

LlavA-next-video-7b \cite{li2024llavanext-strong,zhang2024llavanext-video,liu2024llavanext}: The model is buit on top of LLaVa-NeXT by tuning on a mix of video and image data to achieve better video understanding capabilities. LlavA-next-video-7b-dpo \cite{li2024llavanext-strong,zhang2024llavanext-video,liu2024llavanext}: It aligns the model response with AI feedback using direct preference optimization (DPO), showing a significant performance boost.

4) Models supporting text, image, and video (LM-TIV):

GPT-4o-mini \cite{hurst2024gpt}: The most cost-efficient model from OpenAI that supports text, image and video in the API. Qwen2.5-VL \cite{bai2025qwen2}: The newest flagship model in the Qwen vision-language series showcases remarkable progress in both core capabilities and cutting-edge features. MMAFFLM: Instruction-tune Qwen2.5-VL based on the MMAFFIn dataset. InternVL2.5 \cite{chen2024expanding,gao2024mini,chen2024far,chen2024internvl}: Advanced multilingual and multimodal LM that builds on InternVL 2.0, achieving state-of-the-art performance across diverse benchmarks. InternVL2.5-MPO \cite{wang2024mpo}: Advanced multilingual and multimodal LM that enhances multimodal reasoning based on InternVL2.5 and Mixed Preference Optimization (MPO), achieving superior performance, particularly in Chain-of-Thought tasks.

\section{Results}

Table \ref{tab:resultsSemEval} and Table \ref{tab:resultsall} present the performance of LLMs on all datasets in MMAFFBen. In this section, we provide a step-by-step analysis of the performance of different datasets across various modalities.

\subsection{Performance on Text Datasets}

Due to permission restrictions from SemEval-2018, we report the results on this dataset separately\footnote{It is important to note that despite the limited access to SemEval2018, other data are still sufficient to fully evaluate the affective analysis capabilities of LMs.}, as shown in Table~\ref{tab:resultsSemEval}. Among all models, GPT-4o-mini achieves the best overall performance, followed by EmoLlama-chat-7B~\cite{liu2024emollms}. Notably, EmoLlama-chat-7B performs best on the SemEval\_en subset, likely due to being fine-tuned on its corresponding training data. Although it maintains relatively strong performance in Spanish, its accuracy drops significantly in Arabic, where it is outperformed by both GPT-4o-mini and MMAFFLM-3b/7b.

Table~\ref{tab:resultsall} presents the results on other text-based datasets. MMAFFLM-3b achieves the best overall performance in this broader evaluation, demonstrating its effectiveness across languages and tasks. In contrast, EmoLlama-chat-7B shows limitations in handling multilingual data. One of the most likely reasons is that its post-instruction-tuning training data only contains English data.

Among models that are not fine-tuned for specific tasks, GPT-4o-mini consistently achieves the highest performance. This highlights the strong general capabilities of OpenAI’s model. 
Most other multimodal models perform comparably to text-only models (i.e., LM-T) of similar size on text-only datasets, suggesting that VLMs maintain the text understanding capabilities while extending to additional modalities.

\subsection{Performance on Image Datasets}

On image datasets (Table \ref{tab:resultsall}), GPT-4o-mini achieves the best overall performance, followed by InternVL2.5-8B-MPO. Notably, LM-TI models underperform compared to LM-TIV models, indicating that the integration of video data enhances sentiment analysis performance even on static image tasks.

Compared to GPT-4o-mini and the InternVL2.5 series, the MMAFFLM series demonstrates superior performance on the CFAPS dataset but performs less favorably on other benchmarks. A possible explanation is that MMAFFLM is based on the QwenVL architecture, which is better at Eastern image data, potentially introducing a bias toward specific demographic distributions.

Additionally, results from the InternVL2.5 series suggest that increasing model size leads to significant performance gains, particularly on image datasets—a trend more pronounced than in purely text-based settings.

\subsection{Performance on Video Datasets}

Table~\ref{tab:resultsall} also presents the performance of various models on video-based affective analysis datasets. The MMAFFLM series achieves the best overall performance across most benchmarks. Consistent with observations in the image modality, LM-TV models generally underperform compared to LM-TIV models, suggesting that the combination of image and video training offers complementary benefits for affective understanding.

Interestingly, the MMAFFLM series underperforms GPT-4o-mini and InternVL2.5-8B on the CHSIMS-SI task, a Chinese-language dataset focused on sentiment intensity. This indicates that MMAFFLM has relatively weaker SI-specific capabilities, which is also reflected in the SI-related tasks whthin the image datasets.

As observed in the image modality, increasing model size leads to notable improvements in video affective analysis performance, reinforcing the importance of scale in multimodal understanding.

\subsection{Overall Performance of Trimodal LMs}

The last column in Table~\ref{tab:resultsall} summarizes the overall performance of models that support text, image, and video modalities. Due to licensing restrictions associated with SemEval-2018, only results from other text datasets are included in this comparison. The results indicate that MMAFFLM-7B achieves the best overall performance across modalities, demonstrating that task-specific fine-tuning remains an effective strategy for enhancing multimodal capabilities. Among models without task-specific fine-tuning, GPT-4o-mini, developed by OpenAI, significantly outperforms all other open-source alternatives, highlighting its strong generalization ability across diverse input types.

Additionally, we report the performance of LMs on different languages within the SAMSEMO, MMS, and XAD datasets in Appendix \ref{app:performanceacrosslanguages}. The results demonstrate the multilingual advantage of the MMAFFLM series, followed by GPT-4o-mini.

\section{Conclusion}

In this work, we present MMAFFBen, the first comprehensive, open-source benchmark designed for multilingual and multimodal affective analysis. MMAFFBen integrates text, image, and video modalities across 35 languages, and supports four core affective tasks: sentiment polarity, sentiment intensity, emotion classification, and emotion intensity prediction. We also introduce MMAFFIn, the first instruction-tuning dataset tailored for comprehensive multilingual and multimodal affective analysis. Building on MMAFFIn, we further develop two dedicated MMAFF models specifically designed for affective analysis tasks. We conduct a comprehensive evaluation of several representative LLMs and VLMs, including GPT-4o-mini, offering a systematic comparison of their capabilities in affective understanding. We found that in the field of affective analysis, instruction-tuning remains an effective approach. Among the models without fine-tuning, GPT-4o-mini demonstrated the best overall performance.

Limitations: Due to limitations in computational resources and cost, we only tested open-source models up to 13B in size, as well as GPT-4o-mini. We made every effort to collect representative affective analysis datasets; however, some datasets were not included due to access restrictions.


\bibliographystyle{ACM-Reference-Format}
\bibliography{sample-sigconf}


\begin{thebibliography}{55}


\ifx \showCODEN    \undefined \def \showCODEN     #1{\unskip}     \fi
\ifx \showISBNx    \undefined \def \showISBNx     #1{\unskip}     \fi
\ifx \showISBNxiii \undefined \def \showISBNxiii  #1{\unskip}     \fi
\ifx \showISSN     \undefined \def \showISSN      #1{\unskip}     \fi
\ifx \showLCCN     \undefined \def \showLCCN      #1{\unskip}     \fi
\ifx \shownote     \undefined \def \shownote      #1{#1}          \fi
\ifx \showarticletitle \undefined \def \showarticletitle #1{#1}   \fi
\ifx \showURL      \undefined \def \showURL       {\relax}        \fi
\providecommand\bibfield[2]{#2}
\providecommand\bibinfo[2]{#2}
\providecommand\natexlab[1]{#1}
\providecommand\showeprint[2][]{arXiv:#2}

\bibitem[Acheampong et~al\mbox{.}(2020)]%
        {acheampong2020text}
\bibfield{author}{\bibinfo{person}{Francisca~Adoma Acheampong}, \bibinfo{person}{Chen Wenyu}, {and} \bibinfo{person}{Henry Nunoo-Mensah}.} \bibinfo{year}{2020}\natexlab{}.
\newblock \showarticletitle{Text-based emotion detection: Advances, challenges, and opportunities}.
\newblock \bibinfo{journal}{\emph{Engineering Reports}} \bibinfo{volume}{2}, \bibinfo{number}{7} (\bibinfo{year}{2020}), \bibinfo{pages}{e12189}.
\newblock


\bibitem[Achiam et~al\mbox{.}(2023)]%
        {achiam2023gpt}
\bibfield{author}{\bibinfo{person}{Josh Achiam}, \bibinfo{person}{Steven Adler}, \bibinfo{person}{Sandhini Agarwal}, \bibinfo{person}{Lama Ahmad}, \bibinfo{person}{Ilge Akkaya}, \bibinfo{person}{Florencia~Leoni Aleman}, \bibinfo{person}{Diogo Almeida}, \bibinfo{person}{Janko Altenschmidt}, \bibinfo{person}{Sam Altman}, \bibinfo{person}{Shyamal Anadkat}, {et~al\mbox{.}}} \bibinfo{year}{2023}\natexlab{}.
\newblock \showarticletitle{Gpt-4 technical report}.
\newblock \bibinfo{journal}{\emph{arXiv preprint arXiv:2303.08774}} (\bibinfo{year}{2023}).
\newblock


\bibitem[Ag{\"u}ero-Torales et~al\mbox{.}(2021)]%
        {aguero2021deep}
\bibfield{author}{\bibinfo{person}{Marvin~M Ag{\"u}ero-Torales}, \bibinfo{person}{Jos{\'e} I~Abreu Salas}, {and} \bibinfo{person}{Antonio~G L{\'o}pez-Herrera}.} \bibinfo{year}{2021}\natexlab{}.
\newblock \showarticletitle{Deep learning and multilingual sentiment analysis on social media data: An overview}.
\newblock \bibinfo{journal}{\emph{Applied Soft Computing}}  \bibinfo{volume}{107} (\bibinfo{year}{2021}), \bibinfo{pages}{107373}.
\newblock


\bibitem[Bai et~al\mbox{.}(2025)]%
        {bai2025qwen2}
\bibfield{author}{\bibinfo{person}{Shuai Bai}, \bibinfo{person}{Keqin Chen}, \bibinfo{person}{Xuejing Liu}, \bibinfo{person}{Jialin Wang}, \bibinfo{person}{Wenbin Ge}, \bibinfo{person}{Sibo Song}, \bibinfo{person}{Kai Dang}, \bibinfo{person}{Peng Wang}, \bibinfo{person}{Shijie Wang}, \bibinfo{person}{Jun Tang}, {et~al\mbox{.}}} \bibinfo{year}{2025}\natexlab{}.
\newblock \showarticletitle{Qwen2. 5-VL Technical Report}.
\newblock \bibinfo{journal}{\emph{arXiv preprint arXiv:2502.13923}} (\bibinfo{year}{2025}).
\newblock


\bibitem[Bujnowski et~al\mbox{.}(2024)]%
        {bujnowski2024samsemo}
\bibfield{author}{\bibinfo{person}{Pawe{\l} Bujnowski}, \bibinfo{person}{Bart{\l}omiej Kuzma}, \bibinfo{person}{Bart{\l}omiej Paziewski}, \bibinfo{person}{Jacek Rutkowski}, \bibinfo{person}{Joanna Marhula}, \bibinfo{person}{Zuzanna Bordzicka}, {and} \bibinfo{person}{Piotr Andruszkiewicz}.} \bibinfo{year}{2024}\natexlab{}.
\newblock \showarticletitle{SAMSEMO: New dataset for multilingual and multimodal emotion recognition}. In \bibinfo{booktitle}{\emph{Proc. Interspeech 2024}}. \bibinfo{pages}{2925--2929}.
\newblock


\bibitem[Busso et~al\mbox{.}(2008)]%
        {busso2008iemocap}
\bibfield{author}{\bibinfo{person}{Carlos Busso}, \bibinfo{person}{Murtaza Bulut}, \bibinfo{person}{Chi-Chun Lee}, \bibinfo{person}{Abe Kazemzadeh}, \bibinfo{person}{Emily Mower}, \bibinfo{person}{Samuel Kim}, \bibinfo{person}{Jeannette~N Chang}, \bibinfo{person}{Sungbok Lee}, {and} \bibinfo{person}{Shrikanth~S Narayanan}.} \bibinfo{year}{2008}\natexlab{}.
\newblock \showarticletitle{IEMOCAP: Interactive emotional dyadic motion capture database}.
\newblock \bibinfo{journal}{\emph{Language resources and evaluation}}  \bibinfo{volume}{42} (\bibinfo{year}{2008}), \bibinfo{pages}{335--359}.
\newblock


\bibitem[Chandrasekaran et~al\mbox{.}(2021)]%
        {chandrasekaran2021multimodal}
\bibfield{author}{\bibinfo{person}{Ganesh Chandrasekaran}, \bibinfo{person}{Tu~N Nguyen}, {and} \bibinfo{person}{Jude Hemanth~D}.} \bibinfo{year}{2021}\natexlab{}.
\newblock \showarticletitle{Multimodal sentimental analysis for social media applications: A comprehensive review}.
\newblock \bibinfo{journal}{\emph{Wiley Interdisciplinary Reviews: Data Mining and Knowledge Discovery}} \bibinfo{volume}{11}, \bibinfo{number}{5} (\bibinfo{year}{2021}), \bibinfo{pages}{e1415}.
\newblock


\bibitem[Chen et~al\mbox{.}(2024a)]%
        {chen2024expanding}
\bibfield{author}{\bibinfo{person}{Zhe Chen}, \bibinfo{person}{Weiyun Wang}, \bibinfo{person}{Yue Cao}, \bibinfo{person}{Yangzhou Liu}, \bibinfo{person}{Zhangwei Gao}, \bibinfo{person}{Erfei Cui}, \bibinfo{person}{Jinguo Zhu}, \bibinfo{person}{Shenglong Ye}, \bibinfo{person}{Hao Tian}, \bibinfo{person}{Zhaoyang Liu}, {et~al\mbox{.}}} \bibinfo{year}{2024}\natexlab{a}.
\newblock \showarticletitle{Expanding Performance Boundaries of Open-Source Multimodal Models with Model, Data, and Test-Time Scaling}.
\newblock \bibinfo{journal}{\emph{arXiv preprint arXiv:2412.05271}} (\bibinfo{year}{2024}).
\newblock


\bibitem[Chen et~al\mbox{.}(2024b)]%
        {chen2024far}
\bibfield{author}{\bibinfo{person}{Zhe Chen}, \bibinfo{person}{Weiyun Wang}, \bibinfo{person}{Hao Tian}, \bibinfo{person}{Shenglong Ye}, \bibinfo{person}{Zhangwei Gao}, \bibinfo{person}{Erfei Cui}, \bibinfo{person}{Wenwen Tong}, \bibinfo{person}{Kongzhi Hu}, \bibinfo{person}{Jiapeng Luo}, \bibinfo{person}{Zheng Ma}, {et~al\mbox{.}}} \bibinfo{year}{2024}\natexlab{b}.
\newblock \showarticletitle{How Far Are We to GPT-4V? Closing the Gap to Commercial Multimodal Models with Open-Source Suites}.
\newblock \bibinfo{journal}{\emph{arXiv preprint arXiv:2404.16821}} (\bibinfo{year}{2024}).
\newblock


\bibitem[Chen et~al\mbox{.}(2024c)]%
        {chen2024internvl}
\bibfield{author}{\bibinfo{person}{Zhe Chen}, \bibinfo{person}{Jiannan Wu}, \bibinfo{person}{Wenhai Wang}, \bibinfo{person}{Weijie Su}, \bibinfo{person}{Guo Chen}, \bibinfo{person}{Sen Xing}, \bibinfo{person}{Muyan Zhong}, \bibinfo{person}{Qinglong Zhang}, \bibinfo{person}{Xizhou Zhu}, \bibinfo{person}{Lewei Lu}, {et~al\mbox{.}}} \bibinfo{year}{2024}\natexlab{c}.
\newblock \showarticletitle{Internvl: Scaling up vision foundation models and aligning for generic visual-linguistic tasks}. In \bibinfo{booktitle}{\emph{Proceedings of the IEEE/CVF Conference on Computer Vision and Pattern Recognition}}. \bibinfo{pages}{24185--24198}.
\newblock


\bibitem[Cheng et~al\mbox{.}(2024)]%
        {cheng2024emotion}
\bibfield{author}{\bibinfo{person}{Zebang Cheng}, \bibinfo{person}{Zhi-Qi Cheng}, \bibinfo{person}{Jun-Yan He}, \bibinfo{person}{Jingdong Sun}, \bibinfo{person}{Kai Wang}, \bibinfo{person}{Yuxiang Lin}, \bibinfo{person}{Zheng Lian}, \bibinfo{person}{Xiaojiang Peng}, {and} \bibinfo{person}{Alexander Hauptmann}.} \bibinfo{year}{2024}\natexlab{}.
\newblock \showarticletitle{Emotion-LLaMA: Multimodal Emotion Recognition and Reasoning with Instruction Tuning}.
\newblock \bibinfo{journal}{\emph{arXiv preprint arXiv:2406.11161}} (\bibinfo{year}{2024}).
\newblock


\bibitem[Devika et~al\mbox{.}(2016)]%
        {devika2016sentiment}
\bibfield{author}{\bibinfo{person}{M~D{\textordfeminine} Devika}, \bibinfo{person}{C{\textordfeminine} Sunitha}, {and} \bibinfo{person}{Amal Ganesh}.} \bibinfo{year}{2016}\natexlab{}.
\newblock \showarticletitle{Sentiment analysis: a comparative study on different approaches}.
\newblock \bibinfo{journal}{\emph{Procedia Computer Science}}  \bibinfo{volume}{87} (\bibinfo{year}{2016}), \bibinfo{pages}{44--49}.
\newblock


\bibitem[Dixon-Gordon et~al\mbox{.}(2015)]%
        {dixon2015emotion}
\bibfield{author}{\bibinfo{person}{Katherine~L Dixon-Gordon}, \bibinfo{person}{Amelia Aldao}, {and} \bibinfo{person}{Andres De~Los~Reyes}.} \bibinfo{year}{2015}\natexlab{}.
\newblock \showarticletitle{Emotion regulation in context: Examining the spontaneous use of strategies across emotional intensity and type of emotion}.
\newblock \bibinfo{journal}{\emph{Personality and Individual Differences}}  \bibinfo{volume}{86} (\bibinfo{year}{2015}), \bibinfo{pages}{271--276}.
\newblock


\bibitem[Ekman(1992)]%
        {ekman1992argument}
\bibfield{author}{\bibinfo{person}{Paul Ekman}.} \bibinfo{year}{1992}\natexlab{}.
\newblock \showarticletitle{An argument for basic emotions}.
\newblock \bibinfo{journal}{\emph{Cognition \& emotion}} \bibinfo{volume}{6}, \bibinfo{number}{3-4} (\bibinfo{year}{1992}), \bibinfo{pages}{169--200}.
\newblock


\bibitem[Gaind et~al\mbox{.}(2019)]%
        {gaind2019emotion}
\bibfield{author}{\bibinfo{person}{Bharat Gaind}, \bibinfo{person}{Varun Syal}, {and} \bibinfo{person}{Sneha Padgalwar}.} \bibinfo{year}{2019}\natexlab{}.
\newblock \showarticletitle{Emotion detection and analysis on social media}.
\newblock \bibinfo{journal}{\emph{arXiv preprint arXiv:1901.08458}} (\bibinfo{year}{2019}).
\newblock


\bibitem[Gao et~al\mbox{.}(2024)]%
        {gao2024mini}
\bibfield{author}{\bibinfo{person}{Zhangwei Gao}, \bibinfo{person}{Zhe Chen}, \bibinfo{person}{Erfei Cui}, \bibinfo{person}{Yiming Ren}, \bibinfo{person}{Weiyun Wang}, \bibinfo{person}{Jinguo Zhu}, \bibinfo{person}{Hao Tian}, \bibinfo{person}{Shenglong Ye}, \bibinfo{person}{Junjun He}, \bibinfo{person}{Xizhou Zhu}, {et~al\mbox{.}}} \bibinfo{year}{2024}\natexlab{}.
\newblock \showarticletitle{Mini-internvl: A flexible-transfer pocket multimodal model with 5\% parameters and 90\% performance}.
\newblock \bibinfo{journal}{\emph{arXiv preprint arXiv:2410.16261}} (\bibinfo{year}{2024}).
\newblock


\bibitem[Gong et~al\mbox{.}(2011)]%
        {gong2011revision}
\bibfield{author}{\bibinfo{person}{Xu Gong}, \bibinfo{person}{Yu-Xia Huang}, \bibinfo{person}{Yan Wang}, {and} \bibinfo{person}{Yue-jia Luo}.} \bibinfo{year}{2011}\natexlab{}.
\newblock \showarticletitle{Revision of the Chinese facial affective picture system.}
\newblock \bibinfo{journal}{\emph{Chinese mental health journal}} (\bibinfo{year}{2011}).
\newblock


\bibitem[Hsu et~al\mbox{.}(2018)]%
        {hsu2018emotionlines}
\bibfield{author}{\bibinfo{person}{Chao-Chun Hsu}, \bibinfo{person}{Sheng-Yeh Chen}, \bibinfo{person}{Chuan-Chun Kuo}, \bibinfo{person}{Ting-Hao Huang}, {and} \bibinfo{person}{Lun-Wei Ku}.} \bibinfo{year}{2018}\natexlab{}.
\newblock \showarticletitle{EmotionLines: An Emotion Corpus of Multi-Party Conversations}. In \bibinfo{booktitle}{\emph{Proceedings of the Eleventh International Conference on Language Resources and Evaluation (LREC 2018)}}.
\newblock


\bibitem[Hurst et~al\mbox{.}(2024)]%
        {hurst2024gpt}
\bibfield{author}{\bibinfo{person}{Aaron Hurst}, \bibinfo{person}{Adam Lerer}, \bibinfo{person}{Adam~P Goucher}, \bibinfo{person}{Adam Perelman}, \bibinfo{person}{Aditya Ramesh}, \bibinfo{person}{Aidan Clark}, \bibinfo{person}{AJ Ostrow}, \bibinfo{person}{Akila Welihinda}, \bibinfo{person}{Alan Hayes}, \bibinfo{person}{Alec Radford}, {et~al\mbox{.}}} \bibinfo{year}{2024}\natexlab{}.
\newblock \showarticletitle{Gpt-4o system card}.
\newblock \bibinfo{journal}{\emph{arXiv preprint arXiv:2410.21276}} (\bibinfo{year}{2024}).
\newblock


\bibitem[Hutto and Gilbert(2014)]%
        {hutto2014vader}
\bibfield{author}{\bibinfo{person}{Clayton Hutto} {and} \bibinfo{person}{Eric Gilbert}.} \bibinfo{year}{2014}\natexlab{}.
\newblock \showarticletitle{Vader: A parsimonious rule-based model for sentiment analysis of social media text}. In \bibinfo{booktitle}{\emph{Proceedings of the international AAAI conference on web and social media}}, Vol.~\bibinfo{volume}{8}. \bibinfo{pages}{216--225}.
\newblock


\bibitem[Jiang et~al\mbox{.}(2023)]%
        {jiang2023mistral4}
\bibfield{author}{\bibinfo{person}{Albert~Q Jiang}, \bibinfo{person}{Alexandre Sablayrolles}, \bibinfo{person}{Arthur Mensch}, \bibinfo{person}{Chris Bamford}, \bibinfo{person}{Devendra~Singh Chaplot}, \bibinfo{person}{Diego de~las Casas}, \bibinfo{person}{Florian Bressand}, \bibinfo{person}{Gianna Lengyel}, \bibinfo{person}{Guillaume Lample}, \bibinfo{person}{Lucile Saulnier}, {et~al\mbox{.}}} \bibinfo{year}{2023}\natexlab{}.
\newblock \showarticletitle{Mistral 7B}.
\newblock \bibinfo{journal}{\emph{arXiv preprint arXiv:2310.06825}} (\bibinfo{year}{2023}).
\newblock


\bibitem[Kaggle(2013)]%
        {FER2013}
\bibfield{author}{\bibinfo{person}{Kaggle}.} \bibinfo{year}{2013}\natexlab{}.
\newblock \bibinfo{title}{FER2013 Dataset}.
\newblock
\urldef\tempurl%
\url{https://www.kaggle.com/datasets/msambare/fer2013}
\showURL{%
\tempurl}


\bibitem[Karanikolas et~al\mbox{.}(2023)]%
        {karanikolas2023large}
\bibfield{author}{\bibinfo{person}{Nikitas Karanikolas}, \bibinfo{person}{Eirini Manga}, \bibinfo{person}{Nikoletta Samaridi}, \bibinfo{person}{Eleni Tousidou}, {and} \bibinfo{person}{Michael Vassilakopoulos}.} \bibinfo{year}{2023}\natexlab{}.
\newblock \showarticletitle{Large language models versus natural language understanding and generation}. In \bibinfo{booktitle}{\emph{Proceedings of the 27th Pan-Hellenic Conference on Progress in Computing and Informatics}}. \bibinfo{pages}{278--290}.
\newblock


\bibitem[Kosti et~al\mbox{.}(2020)]%
        {kosti2020context}
\bibfield{author}{\bibinfo{person}{Ronak Kosti}, \bibinfo{person}{Jose~M Alvarez}, \bibinfo{person}{Adria Recasens}, {and} \bibinfo{person}{Agata Lapedriza}.} \bibinfo{year}{2020}\natexlab{}.
\newblock \showarticletitle{Context based emotion recognition using emotic dataset}.
\newblock \bibinfo{journal}{\emph{arXiv preprint arXiv:2003.13401}} (\bibinfo{year}{2020}).
\newblock


\bibitem[Li et~al\mbox{.}(2024b)]%
        {li2024llavanext-strong}
\bibfield{author}{\bibinfo{person}{Bo Li}, \bibinfo{person}{Kaichen Zhang}, \bibinfo{person}{Hao Zhang}, \bibinfo{person}{Dong Guo}, \bibinfo{person}{Renrui Zhang}, \bibinfo{person}{Feng Li}, \bibinfo{person}{Yuanhan Zhang}, \bibinfo{person}{Ziwei Liu}, {and} \bibinfo{person}{Chunyuan Li}.} \bibinfo{year}{2024}\natexlab{b}.
\newblock \bibinfo{title}{LLaVA-NeXT: Stronger LLMs Supercharge Multimodal Capabilities in the Wild}.
\newblock
\urldef\tempurl%
\url{https://llava-vl.github.io/blog/2024-05-10-llava-next-stronger-llms/}
\showURL{%
\tempurl}


\bibitem[Li et~al\mbox{.}(2024a)]%
        {li2024facial}
\bibfield{author}{\bibinfo{person}{Yifan Li}, \bibinfo{person}{Anh Dao}, \bibinfo{person}{Wentao Bao}, \bibinfo{person}{Zhen Tan}, \bibinfo{person}{Tianlong Chen}, \bibinfo{person}{Huan Liu}, {and} \bibinfo{person}{Yu Kong}.} \bibinfo{year}{2024}\natexlab{a}.
\newblock \showarticletitle{Facial Affective Behavior Analysis with Instruction Tuning}.
\newblock \bibinfo{journal}{\emph{European Conference on Computer Vision (ECCV) 2024}} (\bibinfo{year}{2024}).
\newblock


\bibitem[Liao et~al\mbox{.}(2024)]%
        {liao2024vlm2scene}
\bibfield{author}{\bibinfo{person}{Guibiao Liao}, \bibinfo{person}{Jiankun Li}, {and} \bibinfo{person}{Xiaoqing Ye}.} \bibinfo{year}{2024}\natexlab{}.
\newblock \showarticletitle{VLM2Scene: Self-supervised image-text-LiDAR learning with foundation models for autonomous driving scene understanding}. In \bibinfo{booktitle}{\emph{Proceedings of the AAAI Conference on Artificial Intelligence}}, Vol.~\bibinfo{volume}{38}. \bibinfo{pages}{3351--3359}.
\newblock


\bibitem[Lin et~al\mbox{.}(2014)]%
        {lin2014microsoft}
\bibfield{author}{\bibinfo{person}{Tsung-Yi Lin}, \bibinfo{person}{Michael Maire}, \bibinfo{person}{Serge Belongie}, \bibinfo{person}{James Hays}, \bibinfo{person}{Pietro Perona}, \bibinfo{person}{Deva Ramanan}, \bibinfo{person}{Piotr Doll{\'a}r}, {and} \bibinfo{person}{C~Lawrence Zitnick}.} \bibinfo{year}{2014}\natexlab{}.
\newblock \showarticletitle{Microsoft coco: Common objects in context}. In \bibinfo{booktitle}{\emph{Computer Vision--ECCV 2014: 13th European Conference, Zurich, Switzerland, September 6-12, 2014, Proceedings, Part V 13}}. Springer, \bibinfo{pages}{740--755}.
\newblock


\bibitem[link({[n.\,d.]})]%
        {online_shopping_10_cats}
\bibfield{author}{\bibinfo{person}{Github link}.} \bibinfo{year}{[n.\,d.]}\natexlab{}.
\newblock \bibinfo{title}{online\_shopping\_10\_cats}.
\newblock
\urldef\tempurl%
\url{https://github.com/SophonPlus/ChineseNlpCorpus/tree/master/datasets/online\_shopping\_10\_cats}
\showURL{%
\tempurl}


\bibitem[Liu et~al\mbox{.}(2023a)]%
        {liu2023improvedllava}
\bibfield{author}{\bibinfo{person}{Haotian Liu}, \bibinfo{person}{Chunyuan Li}, \bibinfo{person}{Yuheng Li}, {and} \bibinfo{person}{Yong~Jae Lee}.} \bibinfo{year}{2023}\natexlab{a}.
\newblock \bibinfo{title}{Improved Baselines with Visual Instruction Tuning}.
\newblock


\bibitem[Liu et~al\mbox{.}(2024a)]%
        {liu2024llavanext}
\bibfield{author}{\bibinfo{person}{Haotian Liu}, \bibinfo{person}{Chunyuan Li}, \bibinfo{person}{Yuheng Li}, \bibinfo{person}{Bo Li}, \bibinfo{person}{Yuanhan Zhang}, \bibinfo{person}{Sheng Shen}, {and} \bibinfo{person}{Yong~Jae Lee}.} \bibinfo{year}{2024}\natexlab{a}.
\newblock \bibinfo{title}{LLaVA-NeXT: Improved reasoning, OCR, and world knowledge}.
\newblock
\urldef\tempurl%
\url{https://llava-vl.github.io/blog/2024-01-30-llava-next/}
\showURL{%
\tempurl}


\bibitem[Liu et~al\mbox{.}(2023b)]%
        {liu2023llava}
\bibfield{author}{\bibinfo{person}{Haotian Liu}, \bibinfo{person}{Chunyuan Li}, \bibinfo{person}{Qingyang Wu}, {and} \bibinfo{person}{Yong~Jae Lee}.} \bibinfo{year}{2023}\natexlab{b}.
\newblock \bibinfo{title}{Visual Instruction Tuning}.
\newblock


\bibitem[Liu et~al\mbox{.}(2024b)]%
        {liu2024emollms}
\bibfield{author}{\bibinfo{person}{Zhiwei Liu}, \bibinfo{person}{Kailai Yang}, \bibinfo{person}{Qianqian Xie}, \bibinfo{person}{Tianlin Zhang}, {and} \bibinfo{person}{Sophia Ananiadou}.} \bibinfo{year}{2024}\natexlab{b}.
\newblock \showarticletitle{Emollms: A series of emotional large language models and annotation tools for comprehensive affective analysis}. In \bibinfo{booktitle}{\emph{Proceedings of the 30th ACM SIGKDD Conference on Knowledge Discovery and Data Mining}}. \bibinfo{pages}{5487--5496}.
\newblock


\bibitem[Mertens et~al\mbox{.}(2024)]%
        {mertens2024findingemo}
\bibfield{author}{\bibinfo{person}{Laurent Mertens}, \bibinfo{person}{Elahe Yargholi}, \bibinfo{person}{Hans Op~de Beeck}, \bibinfo{person}{Jan Van~den Stock}, {and} \bibinfo{person}{Joost Vennekens}.} \bibinfo{year}{2024}\natexlab{}.
\newblock \showarticletitle{Findingemo: An image dataset for emotion recognition in the wild}.
\newblock \bibinfo{journal}{\emph{Advances in Neural Information Processing Systems}}  \bibinfo{volume}{37} (\bibinfo{year}{2024}), \bibinfo{pages}{4956--4996}.
\newblock


\bibitem[Meta(2024)]%
        {meta2024llama}
\bibfield{author}{\bibinfo{person}{AI Meta}.} \bibinfo{year}{2024}\natexlab{}.
\newblock \showarticletitle{Llama 3.2: Revolutionizing edge AI and vision with open, customizable models}.
\newblock \bibinfo{journal}{\emph{Meta AI Blog. Retrieved December}}  \bibinfo{volume}{20} (\bibinfo{year}{2024}), \bibinfo{pages}{2024}.
\newblock


\bibitem[Mohammad et~al\mbox{.}(2018)]%
        {SemEval2018Task1}
\bibfield{author}{\bibinfo{person}{Saif~M. Mohammad}, \bibinfo{person}{Felipe Bravo-Marquez}, \bibinfo{person}{Mohammad Salameh}, {and} \bibinfo{person}{Svetlana Kiritchenko}.} \bibinfo{year}{2018}\natexlab{}.
\newblock \showarticletitle{SemEval-2018 {T}ask 1: {A}ffect in Tweets}. In \bibinfo{booktitle}{\emph{Proceedings of International Workshop on Semantic Evaluation (SemEval-2018)}}. \bibinfo{address}{New Orleans, LA, USA}.
\newblock


\bibitem[{\"O}hman et~al\mbox{.}(2020)]%
        {ohman2020xed}
\bibfield{author}{\bibinfo{person}{Emily {\"O}hman}, \bibinfo{person}{Marc P{\`a}mies}, \bibinfo{person}{Kaisla Kajava}, {and} \bibinfo{person}{J{\"o}rg Tiedemann}.} \bibinfo{year}{2020}\natexlab{}.
\newblock \showarticletitle{XED: A Multilingual Dataset for Sentiment Analysis and Emotion Detection}. In \bibinfo{booktitle}{\emph{Proceedings of the 28th International Conference on Computational Linguistics}}. \bibinfo{pages}{6542--6552}.
\newblock


\bibitem[Plutchik(1980)]%
        {plutchik1980general}
\bibfield{author}{\bibinfo{person}{Robert Plutchik}.} \bibinfo{year}{1980}\natexlab{}.
\newblock \showarticletitle{A general psychoevolutionary theory of emotion}.
\newblock \bibinfo{journal}{\emph{Emotion: Theory, research, and experience}}  \bibinfo{volume}{1} (\bibinfo{year}{1980}).
\newblock


\bibitem[Poria et~al\mbox{.}(2019)]%
        {poria2019meld}
\bibfield{author}{\bibinfo{person}{Soujanya Poria}, \bibinfo{person}{Devamanyu Hazarika}, \bibinfo{person}{Navonil Majumder}, \bibinfo{person}{Gautam Naik}, \bibinfo{person}{Erik Cambria}, {and} \bibinfo{person}{Rada Mihalcea}.} \bibinfo{year}{2019}\natexlab{}.
\newblock \showarticletitle{MELD: A Multimodal Multi-Party Dataset for Emotion Recognition in Conversations}. In \bibinfo{booktitle}{\emph{Proceedings of the 57th Annual Meeting of the Association for Computational Linguistics}}. \bibinfo{pages}{527--536}.
\newblock


\bibitem[Pramanick et~al\mbox{.}(2024)]%
        {pramanick2024spiqa}
\bibfield{author}{\bibinfo{person}{Shraman Pramanick}, \bibinfo{person}{Rama Chellappa}, {and} \bibinfo{person}{Subhashini Venugopalan}.} \bibinfo{year}{2024}\natexlab{}.
\newblock \showarticletitle{Spiqa: A dataset for multimodal question answering on scientific papers}.
\newblock \bibinfo{journal}{\emph{arXiv preprint arXiv:2407.09413}} (\bibinfo{year}{2024}).
\newblock


\bibitem[Reisenzein and Junge(2024)]%
        {reisenzein2024measuring}
\bibfield{author}{\bibinfo{person}{Rainer Reisenzein} {and} \bibinfo{person}{Martin Junge}.} \bibinfo{year}{2024}\natexlab{}.
\newblock \showarticletitle{Measuring the intensity of emotions}.
\newblock \bibinfo{journal}{\emph{Frontiers in Psychology}}  \bibinfo{volume}{15} (\bibinfo{year}{2024}), \bibinfo{pages}{1437843}.
\newblock


\bibitem[Shaver et~al\mbox{.}(1987)]%
        {shaver1987emotion}
\bibfield{author}{\bibinfo{person}{Phillip Shaver}, \bibinfo{person}{Judith Schwartz}, \bibinfo{person}{Donald Kirson}, {and} \bibinfo{person}{Cary O'connor}.} \bibinfo{year}{1987}\natexlab{}.
\newblock \showarticletitle{Emotion knowledge: further exploration of a prototype approach.}
\newblock \bibinfo{journal}{\emph{Journal of personality and social psychology}} \bibinfo{volume}{52}, \bibinfo{number}{6} (\bibinfo{year}{1987}), \bibinfo{pages}{1061}.
\newblock


\bibitem[Shuai et~al\mbox{.}(2024)]%
        {shuai2024survey}
\bibfield{author}{\bibinfo{person}{Xincheng Shuai}, \bibinfo{person}{Henghui Ding}, \bibinfo{person}{Xingjun Ma}, \bibinfo{person}{Rongcheng Tu}, \bibinfo{person}{Yu-Gang Jiang}, {and} \bibinfo{person}{Dacheng Tao}.} \bibinfo{year}{2024}\natexlab{}.
\newblock \showarticletitle{A survey of multimodal-guided image editing with text-to-image diffusion models}.
\newblock \bibinfo{journal}{\emph{arXiv preprint arXiv:2406.14555}} (\bibinfo{year}{2024}).
\newblock


\bibitem[SMP2020-EWECT(2020)]%
        {SMP2020-EWECT}
\bibfield{author}{\bibinfo{person}{SMP2020-EWECT}.} \bibinfo{year}{2020}\natexlab{}.
\newblock \bibinfo{title}{The Evaluation of Weibo Emotion Classification Technology (SMP2020-EWECT)}.
\newblock
\urldef\tempurl%
\url{https://smp2020ewect.github.io/}
\showURL{%
\tempurl}


\bibitem[Tian et~al\mbox{.}(2018)]%
        {tian2018polarity}
\bibfield{author}{\bibinfo{person}{Leimin Tian}, \bibinfo{person}{Catherine Lai}, {and} \bibinfo{person}{Johanna~D Moore}.} \bibinfo{year}{2018}\natexlab{}.
\newblock \showarticletitle{Polarity and Intensity: the Two Aspects of Sentiment Analysis}. In \bibinfo{booktitle}{\emph{Proceedings of Grand Challenge and Workshop on Human Multimodal Language (Challenge-HML)}}. \bibinfo{pages}{40--47}.
\newblock


\bibitem[Wang et~al\mbox{.}(2024)]%
        {wang2024mpo}
\bibfield{author}{\bibinfo{person}{Weiyun Wang}, \bibinfo{person}{Zhe Chen}, \bibinfo{person}{Wenhai Wang}, \bibinfo{person}{Yue Cao}, \bibinfo{person}{Yangzhou Liu}, \bibinfo{person}{Zhangwei Gao}, \bibinfo{person}{Jinguo Zhu}, \bibinfo{person}{Xizhou Zhu}, \bibinfo{person}{Lewei Lu}, \bibinfo{person}{Yu Qiao}, {and} \bibinfo{person}{Jifeng Dai}.} \bibinfo{year}{2024}\natexlab{}.
\newblock \showarticletitle{Enhancing the Reasoning Ability of Multimodal Large Language Models via Mixed Preference Optimization}.
\newblock \bibinfo{journal}{\emph{arXiv preprint arXiv:2411.10442}} (\bibinfo{year}{2024}).
\newblock


\bibitem[Yang et~al\mbox{.}(2022)]%
        {yang2022emotion}
\bibfield{author}{\bibinfo{person}{Dingkang Yang}, \bibinfo{person}{Shuai Huang}, \bibinfo{person}{Shunli Wang}, \bibinfo{person}{Yang Liu}, \bibinfo{person}{Peng Zhai}, \bibinfo{person}{Liuzhen Su}, \bibinfo{person}{Mingcheng Li}, {and} \bibinfo{person}{Lihua Zhang}.} \bibinfo{year}{2022}\natexlab{}.
\newblock \showarticletitle{Emotion recognition for multiple context awareness}. In \bibinfo{booktitle}{\emph{European conference on computer vision}}. Springer, \bibinfo{pages}{144--162}.
\newblock


\bibitem[Yang et~al\mbox{.}(2025)]%
        {yang2025omni}
\bibfield{author}{\bibinfo{person}{Qize Yang}, \bibinfo{person}{Detao Bai}, \bibinfo{person}{Yi-Xing Peng}, {and} \bibinfo{person}{Xihan Wei}.} \bibinfo{year}{2025}\natexlab{}.
\newblock \showarticletitle{Omni-Emotion: Extending Video MLLM with Detailed Face and Audio Modeling for Multimodal Emotion Analysis}.
\newblock \bibinfo{journal}{\emph{arXiv preprint arXiv:2501.09502}} (\bibinfo{year}{2025}).
\newblock


\bibitem[Yang et~al\mbox{.}(2024)]%
        {yang2024emollm}
\bibfield{author}{\bibinfo{person}{Qu Yang}, \bibinfo{person}{Mang Ye}, {and} \bibinfo{person}{Bo Du}.} \bibinfo{year}{2024}\natexlab{}.
\newblock \showarticletitle{Emollm: Multimodal emotional understanding meets large language models}.
\newblock \bibinfo{journal}{\emph{arXiv preprint arXiv:2406.16442}} (\bibinfo{year}{2024}).
\newblock


\bibitem[Yu et~al\mbox{.}(2020)]%
        {yu2020ch}
\bibfield{author}{\bibinfo{person}{Wenmeng Yu}, \bibinfo{person}{Hua Xu}, \bibinfo{person}{Fanyang Meng}, \bibinfo{person}{Yilin Zhu}, \bibinfo{person}{Yixiao Ma}, \bibinfo{person}{Jiele Wu}, \bibinfo{person}{Jiyun Zou}, {and} \bibinfo{person}{Kaicheng Yang}.} \bibinfo{year}{2020}\natexlab{}.
\newblock \showarticletitle{Ch-sims: A chinese multimodal sentiment analysis dataset with fine-grained annotation of modality}. In \bibinfo{booktitle}{\emph{Proceedings of the 58th annual meeting of the association for computational linguistics}}. \bibinfo{pages}{3718--3727}.
\newblock


\bibitem[Zadeh et~al\mbox{.}(2018)]%
        {zadeh2018multimodal}
\bibfield{author}{\bibinfo{person}{AmirAli~Bagher Zadeh}, \bibinfo{person}{Paul~Pu Liang}, \bibinfo{person}{Soujanya Poria}, \bibinfo{person}{Erik Cambria}, {and} \bibinfo{person}{Louis-Philippe Morency}.} \bibinfo{year}{2018}\natexlab{}.
\newblock \showarticletitle{Multimodal language analysis in the wild: Cmu-mosei dataset and interpretable dynamic fusion graph}. In \bibinfo{booktitle}{\emph{Proceedings of the 56th Annual Meeting of the Association for Computational Linguistics (Volume 1: Long Papers)}}. \bibinfo{pages}{2236--2246}.
\newblock


\bibitem[Zhang et~al\mbox{.}(2024)]%
        {zhang2024llavanext-video}
\bibfield{author}{\bibinfo{person}{Yuanhan Zhang}, \bibinfo{person}{Bo Li}, \bibinfo{person}{haotian Liu}, \bibinfo{person}{Yong~jae Lee}, \bibinfo{person}{Liangke Gui}, \bibinfo{person}{Di Fu}, \bibinfo{person}{Jiashi Feng}, \bibinfo{person}{Ziwei Liu}, {and} \bibinfo{person}{Chunyuan Li}.} \bibinfo{year}{2024}\natexlab{}.
\newblock \bibinfo{title}{LLaVA-NeXT: A Strong Zero-shot Video Understanding Model}.
\newblock
\urldef\tempurl%
\url{https://llava-vl.github.io/blog/2024-04-30-llava-next-video/}
\showURL{%
\tempurl}


\bibitem[Zheng et~al\mbox{.}(2024)]%
        {zheng2024llamafactory}
\bibfield{author}{\bibinfo{person}{Yaowei Zheng}, \bibinfo{person}{Richong Zhang}, \bibinfo{person}{Junhao Zhang}, \bibinfo{person}{Yanhan Ye}, \bibinfo{person}{Zheyan Luo}, \bibinfo{person}{Zhangchi Feng}, {and} \bibinfo{person}{Yongqiang Ma}.} \bibinfo{year}{2024}\natexlab{}.
\newblock \showarticletitle{LlamaFactory: Unified Efficient Fine-Tuning of 100+ Language Models}. In \bibinfo{booktitle}{\emph{Proceedings of the 62nd Annual Meeting of the Association for Computational Linguistics (Volume 3: System Demonstrations)}}. \bibinfo{publisher}{Association for Computational Linguistics}, \bibinfo{address}{Bangkok, Thailand}.
\newblock
\urldef\tempurl%
\url{http://arxiv.org/abs/2403.13372}
\showURL{%
\tempurl}


\bibitem[Zhou et~al\mbox{.}(2019)]%
        {zhou2019semantic}
\bibfield{author}{\bibinfo{person}{Bolei Zhou}, \bibinfo{person}{Hang Zhao}, \bibinfo{person}{Xavier Puig}, \bibinfo{person}{Tete Xiao}, \bibinfo{person}{Sanja Fidler}, \bibinfo{person}{Adela Barriuso}, {and} \bibinfo{person}{Antonio Torralba}.} \bibinfo{year}{2019}\natexlab{}.
\newblock \showarticletitle{Semantic understanding of scenes through the ade20k dataset}.
\newblock \bibinfo{journal}{\emph{International Journal of Computer Vision}}  \bibinfo{volume}{127} (\bibinfo{year}{2019}), \bibinfo{pages}{302--321}.
\newblock


\bibitem[Łukasz Augustyniak et~al\mbox{.}(2023)]%
        {augustyniak2023massively}
\bibfield{author}{\bibinfo{person}{Łukasz Augustyniak}, \bibinfo{person}{Szymon Woźniak}, \bibinfo{person}{Marcin Gruza}, \bibinfo{person}{Piotr Gramacki}, \bibinfo{person}{Krzysztof Rajda}, \bibinfo{person}{Mikołaj Morzy}, {and} \bibinfo{person}{Tomasz Kajdanowicz}.} \bibinfo{year}{2023}\natexlab{}.
\newblock \bibinfo{title}{Massively Multilingual Corpus of Sentiment Datasets and Multi-faceted Sentiment Classification Benchmark}.
\newblock
\showeprint[arxiv]{2306.07902}~[cs.CL]


\end{thebibliography}


\appendix

\section{Performance Across Languages on Multilingual Datasets (Table \ref{tab:multilingualSAMSEMO} for SAMSEMO, Table \ref{tab:multilingualMMS} for MMS, Table \ref{tab:multilingualXED} for XED.) \label{app:performanceacrosslanguages}}

The results demonstrate the multilingual advantage of the MMAFFLM series, followed by GPT-4o-mini. Moreover, we observe that most models perform better on English compared to other languages. A possible reason is that the pretraining corpora of most foundation models are predominantly English-centric. Therefore, advancing multilingual research in LLMs and VLMs is both necessary and meaningful.

\begin{table}[hb]
\caption{Multi-lingual macro F1 on SAMSEMO.}
  \label{tab:multilingualSAMSEMO}
\footnotesize
\begin{tabular}{lccccc}
\hline
Models                 & de            & en            & es            & ko            & pl            \\ \hline
llava-nextvideo-7b     & 30.1          & 32.5          & 25.5          & 26.2          & 11.1          \\
llava-nextvideo-7b-dpo & 31.8          & 30.1          & 27.2          & 24.7          & 10.6          \\ \hline
qwen2.5-vl-3b          & 32.1          & 33.6          & 28.8          & 28.9          & 28.4          \\
qwen2.5-vl-7b          & 29.1          & 30.3          & 28.8          & 41.3          & 27.4          \\
InternVL2.5-1B         & 20.6          & 23.3          & 21.5          & 22.0          & 23.2          \\
InternVL2.5-2B         & 25.1          & 28.2          & 26.8          & 39.3          & 27.6          \\
InternVL2.5-8B         & 36.7          & 38.6          & 32.1          & {\ul 49.5}    & 31.8          \\
InternVL2.5-1B-MPO     & 33.2          & 27.9          & 29.1          & 29.5          & 26.4          \\
InternVL2.5-2B-MPO     & 34.1          & 34.2          & 31.6          & 31.3          & 32.6          \\
InternVL2.5-8B-MPO     & {\ul 37.9}    & 39.2          & 35.3          & 37.6          & {\ul 39.7}    \\
GPT-4o-mini            & \textbf{42.0} & 42.3          & 35.2          & 42.8          & 37.3          \\
MMAFFLM-3b             & 36.1          & {\ul 52.2}    & {\ul 36.1}    & 46.7          & \textbf{48.1} \\
MMAFFLM-7b             & 37.7          & \textbf{52.7} & \textbf{38.5} & \textbf{60.5} & 36.2          \\ \hline
\end{tabular}
\end{table}

\begin{table*}[htb]
\caption{Multi-lingual macro F1 on MMS.}
  \label{tab:multilingualMMS}
\resizebox{\textwidth}{!}{
\begin{tabular}{lccccccccccccccccccccccccccc}
\hline
Models                 & ar            & bg            & bs            & cs            & de            & en            & es            & fa            & fr            & he            & hi            & hr            & hu            & it            & ja            & lv            & pl            & pt            & ru            & sk            & sl            & sq            & sr            & sv            & th            & ur            & zh            \\ \hline
EmoLlama-chat-7B       & 37.9          & 44.9          & 46.0          & 52.6          & 64.7          & 63.3          & 59.9          & 29.2          & 65.7          & 29.2          & 48.3          & 47.2          & 40.5          & 50.6          & 39.3          & 38.9          & 48.9          & 45.0          & 55.3          & 39.1          & 46.6          & 25.9          & 44.6          & 46.5          & 37.2          & 30.3          & 50.9          \\
Llama3.2-1b-instruct   & 18.1          & 28.9          & 31.7          & 31.7          & 31.9          & 31.9          & 36.9          & 31.1          & 32.8          & 17.2          & 25.9          & 30.9          & 27.4          & 37.6          & 31.3          & 29.7          & 32.3          & 27.5          & 30.7          & 30.8          & 33.1          & 21.0          & 31.5          & 36.4          & 25.1          & 23.7          & 31.4          \\
Llama3.2-3b-instruct   & 21.8          & 34.9          & 28.7          & 35.7          & 33.7          & 34.6          & 33.2          & 46.1          & 31.6          & 30.3          & 28.2          & 26.3          & 31.5          & 39.5          & 36.2          & 34.0          & 39.9          & 30.5          & 37.0          & 32.0          & 35.2          & 19.9          & 30.8          & 35.5          & 31.9          & 25.3          & 25.5          \\
Mistral-7b-instruct    & 26.0          & 29.0          & 29.4          & 33.6          & 37.6          & 44.1          & 33.5          & 29.9          & 36.8          & 34.0          & 34.0          & 27.7          & 24.9          & 37.7          & 50.9          & 26.2          & 33.0          & 27.7          & 36.5          & 25.8          & 29.9          & 21.4          & 30.9          & 35.7          & 32.7          & 27.0          & 42.0          \\ \hline
llama3.2-11B           & 28.3          & 36.1          & 37.0          & 37.6          & 37.9          & 51.3          & 42.2          & 41.5          & 50.4          & 36.1          & 34.0          & 35.9          & 36.5          & 46.3          & 51.0          & 34.1          & 38.8          & 31.2          & 40.4          & 35.2          & 34.4          & 24.4          & 32.5          & 40.1          & 34.7          & 28.6          & 33.0          \\
llava-1.5-7b           & 26.2          & 29.6          & 32.5          & 33.2          & 37.3          & 34.1          & 37.0          & 29.1          & 46.1          & 29.8          & 33.2          & 30.4          & 27.8          & 40.1          & 36.6          & 26.6          & 36.5          & 28.0          & 38.5          & 30.6          & 30.4          & 20.6          & 33.9          & 36.4          & 22.3          & 24.9          & 31.6          \\
llava-1.5-13b          & 27.2          & 35.1          & 33.8          & 37.3          & 40.9          & 38.3          & 39.4          & 35.2          & 41.4          & 35.1          & 33.7          & 34.4          & 33.3          & 42.5          & 44.2          & 24.1          & 40.9          & 30.2          & 39.1          & 38.1          & 34.2          & 23.6          & 30.9          & 40.5          & 31.9          & 28.6          & 40.3          \\ \hline
llava-nextvideo-7b     & 26.3          & 29.1          & 31.0          & 35.5          & 32.4          & 36.9          & 36.1          & 26.1          & 35.2          & 21.5          & 29.6          & 30.8          & 30.4          & 29.8          & 33.5          & 32.5          & 33.4          & 24.6          & 33.0          & 31.0          & 25.2          & 25.2          & 22.3          & 28.7          & 23.5          & 25.7          & 37.3          \\
llava-nextvideo-7b-dpo & 27.1          & 29.7          & 29.3          & 38.2          & 44.0          & 37.6          & 35.3          & 29.3          & 45.2          & 25.4          & 31.9          & 34.5          & 42.9          & 28.2          & 36.2          & 33.9          & 40.8          & 34.5          & 34.7          & 32.8          & 27.3          & 26.1          & 22.5          & 31.0          & 22.9          & 27.7          & 38.5          \\ \hline
qwen2.5-vl-3b          & 31.7          & 30.7          & 30.5          & 32.9          & 39.4          & 45.2          & 39.5          & 32.9          & 39.0          & 35.4          & 34.2          & 30.1          & 28.8          & 44.2          & 42.1          & 28.8          & 38.1          & 31.7          & 38.9          & 28.2          & 30.1          & 20.2          & 33.2          & 37.1          & 38.6          & 24.6          & 36.1          \\
qwen2.5-vl-7b          & 26.1          & 29.3          & 24.9          & 30.8          & 35.6          & 38.3          & 35.4          & 28.8          & 33.3          & 32.3          & 30.0          & 26.9          & 26.5          & 34.8          & 36.4          & 26.6          & 29.8          & 23.2          & 36.1          & 28.3          & 26.7          & 18.0          & 26.2          & 32.8          & 30.7          & 20.7          & 33.4          \\
InternVL2.5-1B         & 32.1          & 29.7          & 26.3          & 31.5          & 38.8          & 40.3          & 36.8          & 31.2          & 54.6          & 30.4          & 30.9          & 25.0          & 24.6          & 39.2          & 41.5          & 23.8          & 34.5          & 32.6          & 35.6          & 24.3          & 28.1          & 18.5          & 25.8          & 29.0          & 40.9          & 26.3          & 38.9          \\
InternVL2.5-2B         & 29.5          & 27.4          & 28.0          & 33.7          & 50.7          & 53.8          & 34.1          & 32.0          & 52.8          & 31.3          & 33.0          & 27.4          & 32.9          & 35.2          & 56.8          & 26.4          & 35.5          & 29.8          & 36.9          & 27.9          & 28.6          & 17.9          & 28.8          & 29.3          & 24.9          & 26.7          & 46.3          \\
InternVL2.5-8B         & 30.2          & 29.7          & 34.1          & 34.0          & 51.6          & 62.6          & 46.6          & 27.2          & 37.2          & 40.2          & 34.0          & 30.9          & 29.7          & 41.1          & 55.4          & 27.0          & 34.8          & 28.8          & 51.0          & 33.9          & 30.4          & 23.8          & 32.5          & 35.4          & 34.7          & 30.8          & 47.4          \\
InternVL2.5-1B-MPO     & 28.7          & 25.7          & 25.7          & 36.5          & 45.6          & 34.6          & 35.9          & 29.0          & 49.1          & 35.7          & 28.7          & 22.7          & 24.0          & 36.4          & 53.3          & 23.2          & 30.0          & 41.2          & 31.6          & 22.3          & 24.9          & 20.0          & 27.5          & 29.8          & 43.7          & 26.0          & 33.3          \\
InternVL2.5-2B-MPO     & 36.9          & 36.7          & 25.9          & 39.4          & 43.7          & 44.1          & 41.6          & 30.2          & 46.5          & 32.9          & 30.4          & 26.9          & 31.0          & 44.3          & 48.2          & 21.6          & 32.1          & 27.9          & 43.6          & 26.3          & 33.8          & 16.1          & 24.3          & 38.5          & 22.1          & 24.8          & 37.6          \\
InternVL2.5-8B-MPO     & 27.5          & 26.6          & 28.8          & 43.2          & 43.8          & 42.0          & 44.8          & 27.2          & 35.6          & 51.1          & 44.1          & 27.2          & 34.0          & 49.2          & 47.5          & 27.8          & 32.9          & 38.0          & 44.3          & 38.3          & 37.8          & 19.2          & 37.0          & 42.2          & 38.0          & 27.6          & 40.3          \\
GPT-4o-mini            & 52.6          & \textbf{63.3} & \textbf{62.1} & {\ul 60.5}    & \textbf{69.6} & \textbf{74.0} & \textbf{65.2} & \textbf{75.5} & \textbf{69.7} & \textbf{57.9} & \textbf{61.5} & \textbf{55.2} & \textbf{60.4} & \textbf{76.2} & \textbf{69.7} & \textbf{67.3} & \textbf{66.5} & \textbf{48.2} & \textbf{65.2} & \textbf{58.7} & \textbf{61.2} & \textbf{45.9} & \textbf{56.3} & \textbf{62.6} & \textbf{66.8} & \textbf{47.7} & 59.5 \\
MMAFFLM-3b             & {\ul 57.7}    & {\ul 55.7}    & {\ul 52.0}    & \textbf{61.3} & {\ul 66.9}    & {\ul 72.3}    & {\ul 58.1}    & {\ul 56.5}    & 64.5          & {\ul 55.5}    & {\ul 52.6}    & 52.9          & {\ul 53.6}    & {\ul 62.4}    & 61.1          & {\ul 49.0}    & {\ul 60.5}    & 44.2          & 59.5          & 49.7          & 49.1          & 38.7          & 48.1          & 52.6          & {\ul 44.6}    & 44.6          & \textbf{64.0}    \\
MMAFFLM-7b             & \textbf{60.7} & 53.6          & 47.6          & 59.7          & 63.7          & 69.9          & 57.6          & 51.3          & {\ul 67.0}    & 50.1          & 52.2          & {\ul 55.0}    & 48.6          & {\ul 62.4}    & {\ul 61.6}    & 44.5          & 59.8          & {\ul 44.9}    & {\ul 59.7}    & {\ul 52.9}    & {\ul 51.9}    & {\ul 44.4}    & {\ul 49.2}    & {\ul 53.4}    & 40.9          & {\ul 45.1}    & {\ul 62.7}          \\ \hline
\end{tabular}
}
\end{table*}

\begin{table*}[htb]
\caption{Multi-lingual macro F1 on XED.}
  \label{tab:multilingualXED}
\resizebox{\textwidth}{!}{
\begin{tabular}{lccccccccccccccccccccccccc}
\hline
Model                  & ar            & bg            & br            & bs            & cs            & de            & el            & en            & es            & fi            & fr            & he            & hr            & hu            & it            & nl            & no            & pl            & pt            & ro            & ru            & sl            & sr            & sv            & tr            \\ \hline
EmoLlama-chat-7B       & 14.8          & 20.3          & 23.0          & 14.4          & 16.8          & 22.0          & 18.6          & 38.6          & 23.5          & 18.7          & 24.6          & 16.8          & 16.7          & 15.7          & 23.7          & 20.3          & 19.6          & 21.2          & 24.6          & 20.8          & 22.4          & 14.9          & 15.0          & 19.1          & 9.0           \\
Llama3.2-1b-instruct   & 17.8          & 19.7          & 20.0          & 18.9          & 18.5          & 18.6          & 17.8          & 30.3          & 21.1          & 15.5          & 18.3          & 14.2          & 18.6          & 19.0          & 21.2          & 16.1          & 18.2          & 20.1          & 19.2          & 18.6          & 15.0          & 17.0          & 16.6          & 16.2          & 12.1          \\
Llama3.2-3b-instruct   & 9.8           & 6.7           & 6.4           & 7.7           & 7.5           & 7.3           & 9.8           & 19.2          & 9.0           & 8.0           & 9.5           & 11.6          & 9.4           & 7.7           & 10.4          & 9.0           & 8.5           & 7.2           & 4.1           & 10.4          & 8.8           & 7.7           & 12.5          & 5.8           & 7.1           \\
Mistral-7b-instruct    & 22.2          & 22.3          & 24.1          & 21.6          & 22.5          & 25.5          & 19.4          & 31.7          & 25.2          & 22.6          & 23.1          & 22.2          & 22.1          & {\ul 22.4}    & 23.2          & 21.3          & 22.8          & 24.5          & 24.7          & {\ul 24.1}    & 25.1          & 22.7          & 22.1          & 23.9          & 16.7          \\ \hline
llama3.2-11B           & 18.6          & 21.2          & 19.0          & 18.1          & 19.1          & 20.4          & 23.9          & 31.7          & 23.4          & 19.3          & 20.3          & 21.5          & 17.7          & 15.6          & 21.5          & 19.2          & 17.1          & 19.3          & 20.6          & 18.1          & 22.1          & 14.3          & 16.4          & 19.7          & 16.0          \\
llava-1.5-7b           & 21.1          & 22.7          & 23.8          & 20.3          & 21.7          & 24.9          & 19.5          & 33.0          & {\ul 25.4}    & 23.6          & 24.8          & 22.0          & 21.9          & 20.8          & 24.2          & 22.5          & 23.2          & 20.8          & {\ul 25.2}    & 22.8          & 24.8          & 20.3          & 20.8          & 25.2          & 17.3          \\
llava-1.5-13b          & 17.9          & 24.2          & {\ul 24.4}    & 18.3          & 20.8          & {\ul 26.2}    & 18.9          & 30.9          & 25.3          & 24.3          & 25.0          & 21.5          & 19.5          & 19.3          & 21.7          & 20.9          & {\ul 24.8}    & 21.0          & 21.2          & 21.8          & 25.7          & 20.0          & 19.4          & {\ul 25.4}    & 14.6          \\ \hline
llava-nextvideo-7b     & 19.5          & 18.3          & 18.2          & 15.3          & 18.0          & 18.1          & 16.8          & 26.3          & 20.4          & 16.7          & 18.5          & 16.8          & 14.9          & 17.8          & 18.2          & 16.6          & 17.1          & 16.2          & 16.4          & 17.7          & 22.1          & 15.0          & 16.5          & 17.9          & 13.6          \\
llava-nextvideo-7b-dpo & 20.2          & 17.7          & 18.8          & 17.4          & 18.8          & 19.4          & 18.8          & 28.2          & 21.4          & 16.3          & 19.9          & 16.9          & 19.0          & 17.8          & 20.7          & 17.8          & 17.1          & 16.6          & 22.0          & 17.7          & 20.6          & 17.4          & 15.3          & 19.2          & 15.6          \\ \hline
qwen2.5-vl-3b          & 14.0          & 11.0          & 14.7          & 10.2          & 10.2          & 12.2          & 13.1          & 23.3          & 14.9          & 10.9          & 17.4          & 14.7          & 10.5          & 11.3          & 17.1          & 12.6          & 13.5          & 14.7          & 15.1          & 14.2          & 17.8          & 8.6           & 10.3          & 13.7          & 9.9           \\
qwen2.5-vl-7b          & 12.8          & 13.2          & 12.8          & 9.1           & 9.5           & 17.1          & 12.1          & 26.4          & 16.0          & 12.0          & 14.8          & 14.7          & 9.0           & 7.2           & 14.3          & 11.8          & 10.1          & 10.3          & 12.2          & 11.8          & 18.4          & 8.0           & 10.0          & 10.2          & 8.9           \\
InternVL2.5-1B         & 21.2          & 19.0          & 21.2          & 16.5          & 17.7          & 20.4          & 12.9          & 17.7          & 17.5          & 15.3          & 18.4          & 20.4          & 16.3          & 18.2          & 19.6          & 19.9          & 16.1          & 17.5          & 21.7          & 20.4          & 20.8          & 15.4          & 15.8          & 19.1          & 14.4          \\
InternVL2.5-2B         & 4.4           & 4.8           & 5.8           & 2.7           & 4.5           & 5.3           & 5.0           & 12.9          & 5.2           & 5.9           & 5.5           & 5.5           & 1.2           & 1.6           & 3.4           & 3.2           & 2.5           & 5.2           & 5.0           & 4.8           & 6.1           & 2.4           & 1.2           & 4.0           & 2.6           \\
InternVL2.5-8B         & 7.4           & 8.0           & 11.4          & 6.6           & 6.9           & 10.1          & 7.7           & 24.0          & 12.6          & 11.0          & 11.9          & 7.2           & 6.9           & 6.7           & 9.1           & 8.0           & 9.6           & 6.7           & 15.1          & 10.5          & 10.8          & 5.3           & 3.5           & 10.2          & 9.0           \\
InternVL2.5-1B-MPO     & 24.1          & 22.2          & 22.4          & 21.3          & 21.5          & 21.7          & 19.5          & 25.1          & 22.3          & 16.6          & 20.5          & {\ul 23.5}    & 19.1          & 18.3          & 21.7          & 21.6          & 19.5          & 19.9          & 23.7          & 20.3          & 23.9          & 18.8          & 19.8          & 19.8          & 16.0          \\
InternVL2.5-2B-MPO     & 11.2          & 11.8          & 11.1          & 10.7          & 10.0          & 13.9          & 9.7           & 15.1          & 9.3           & 11.5          & 10.8          & 12.5          & 10.5          & 9.3           & 8.8           & 9.3           & 8.5           & 11.4          & 14.1          & 10.1          & 17.0          & 7.6           & 9.1           & 10.6          & 9.2           \\
InternVL2.5-8B-MPO     & 7.9           & 9.5           & 13.8          & 9.8           & 12.2          & 13.9          & 9.8           & 27.3          & 17.5          & 14.0          & 14.3          & 6.9           & 11.2          & 10.9          & 14.1          & 8.4           & 13.2          & 9.5           & 17.9          & 12.6          & 11.5          & 8.0           & 5.0           & 14.0          & 11.1          \\
GPT-4o-mini            & 18.6          & 21.7          & 18.0          & 19.6          & 21.1          & 22.7          & 19.8          & {\ul 34.1}    & 23.0          & 24.0          & 19.9          & 22.9          & 18.6          & 18.3          & 20.6          & 19.3          & 21.9          & 20.0          & 21.8          & 21.0          & 20.6          & 16.7          & 21.9          & 22.5          & 17.1          \\
MMAFFLM-3b             & \textbf{27.2} & {\ul 26.7}    & \textbf{26.0} & {\ul 23.3}    & \textbf{37.3} & \textbf{28.5} & \textbf{35.4} & \textbf{36.8} & \textbf{29.0} & \textbf{37.7} & \textbf{26.7} & \textbf{35.0} & {\ul 28.4}    & \textbf{31.9} & {\ul 26.7}    & \textbf{26.3} & \textbf{26.6} & {\ul 26.8}    & \textbf{28.4} & \textbf{25.7} & {\ul 29.6}    & {\ul 22.9}    & {\ul 23.6}    & \textbf{27.1} & {\ul 20.2}    \\
MMAFFLM-7b             & {\ul 24.2}    & \textbf{33.7} & 21.6          & \textbf{31.2} & {\ul 24.3}    & 25.4          & {\ul 31.0}    & 29.9          & 24.6          & {\ul 32.4}    & {\ul 25.5}    & 23.4          & \textbf{30.1} & 19.1          & \textbf{36.1} & {\ul 23.2}    & 23.2          & \textbf{32.3} & 23.3          & 20.3          & \textbf{37.7} & \textbf{29.0} & \textbf{31.1} & 24.1          & \textbf{32.2} \\ \hline
\end{tabular}
}
\end{table*}

\section{Prompt descriptions for each dataset and each task (Table \ref{tab:promptdescriptions})\label{app:promptdescription}}

\begin{table*}[]
\footnotesize
\caption{Prompt descriptions for each dataset and each task}
  \label{tab:promptdescriptions}
\begin{tabular}{lp{15cm}}
\hline
{\color[HTML]{212121} Dataset}                      & Prompt                                                                                                                                                                                                                                                                                                                                                                                                                                                                                                                                                                                                                                                                                                                                                                                                                                                                                                                                                                                                                                                                                                                                                                                                                                                                                                                                                                                                                                                                                                                                                                                                                                                                                                                                                                                                                                                                                                                                                                                                                                                                                                                                                                                                                                                                                                                                                                                                                                                                                                                                           \\ \hline
SemEva2018-EI   \cite{SemEval2018Task1}               & Task: Assess the   magnitude of emotion E in the text using a real number between 0 and 1, where   0 denotes the least intensity and 1 denotes the most intensity. Text: {raw   text}. Emotion E: anger. Intensity Score:                                                                                                                                                                                                                                                                                                                                                                                                                                                                                                                                                                                                                                                                                                                                                                                                                                                                                                                                                                                                                                                                                                                                                                                                                                                                                                                                                                                                                                                                                                                                                                                                                                                                                                                                                                                                                                                                                                                                                                                                                                                                                                                                                                                                                                                                                                                        \\ \hline
SemEva2018-EC (multi)   \cite{SemEval2018Task1}   & \begin{tabular}[c]{@{}l@{}}Task: Determine the   text's emotional tone based on 11 categories: 1. joy, 2. sadness, 3. anger,   4. fear, 5. surprise, 6. disgust, \\7. anticipation, 8. love, 9. optimism, 10.   pessimism, and 11. trust. \\      You have two options:\\      (a) Select one or more of the 11 emotions that represent the emotional   state depicted in the text. \\      (b) Choose ‘0. neutral’ if the image does not express any emotion.\\      Please provide your selection in the following format: {number}.   {emotion}.\\      Text: {raw text}\end{tabular}                                                                                                                                                                                                                                                                                                                                                                                                                                                                                                                                                                                                                                                                                                                                                                                                                                                                                                                                                                                                                                                                                                                                                                                                                                                                                                                                                                                                                                                                                                                                                                                                                                                                                                                                                                                                                                                                                                                                                           \\ \hline
SemEva2018-SP (7)   \cite{SemEval2018Task1}       & Task: Classify the   text into one of seven ordinal classes, corresponding to various levels of   positive and negative sentiment intensity, that best represents the mental   state of the publisher. 3: very positive mental state can be inferred. 2:   moderately positive mental state can be inferred. 1: slightly positive mental   state can be inferred. 0: neutral or mixed mental state can be inferred. -1:   slightly negative mental state can be inferred. -2: moderately negative   mental state can be inferred. -3: very negative mental state can be inferred   Text: {raw text}. Intensity Class:                                                                                                                                                                                                                                                                                                                                                                                                                                                                                                                                                                                                                                                                                                                                                                                                                                                                                                                                                                                                                                                                                                                                                                                                                                                                                                                                                                                                                                                                                                                                                                                                                                                                                                                                                                                                                                                                                                                            \\ \hline
SemEva2018-SI   \cite{SemEval2018Task1}      & Task: Determine the   sentiment (valence) intensity of the publisher's mental state on a scale of   -1 (most negative) to 1 (most positive). Text: {raw text}. Intensity   Score:                                                                                                                                                                                                                                                                                                                                                                                                                                                                                                                                                                                                                                                                                                                                                                                                                                                                                                                                                                                                                                                                                                                                                                                                                                                                                                                                                                                                                                                                                                                                                                                                                                                                                                                                                                                                                                                                                                                                                                                                                                                                                                                                                                                                                                                                                                                                                                \\ \hline
EWECT-usual-EC (single)                          & \begin{tabular}[c]{@{}l@{}}Task: Determine the   emotion of the text based on five discrete categories: 1. happiness, 2.   sadness, 3. anger, 4. fear, and \\5. surprise. \\      You have two options:\\      (a) Select one emotion that best represents the emotional state conveyed in   the text.\\      (b) Choose ‘0. neutral’ if the text does not express any emotion.\\      Please provide your selection in the following format: {number}.   {emotion}.\\      Text: {raw text}\end{tabular}                                                                                                                                                                                                                                                                                                                                                                                                                                                                                                                                                                                                                                                                                                                                                                                                                                                                                                                                                                                                                                                                                                                                                                                                                                                                                                                                                                                                                                                                                                                                                                                                                                                                                                                                                                                                                                                                                                                                                                                                                                          \\ \hline
EWECT-virus-EC (single)                        & \begin{tabular}[c]{@{}l@{}}Task: Determine the   emotion of the text based on five discrete categories: 1. happiness, 2.   sadness, 3. anger, 4. fear, and \\5. surprise. \\      You have two options:\\      (a) Select one emotion that best represents the emotional state conveyed in   the text.\\      (b) Choose ‘0. neutral’ if the text does not express any emotion.\\      Please provide your selection in the following format: {number}.   {emotion}.\\      Text: {raw text}\end{tabular}                                                                                                                                                                                                                                                                                                                                                                                                                                                                                                                                                                                                                                                                                                                                                                                                                                                                                                                                                                                                                                                                                                                                                                                                                                                                                                                                                                                                                                                                                                                                                                                                                                                                                                                                                                                                                                                                                                                                                                                                                                          \\ \hline
Onlineshopping-SP (2)                           & \begin{tabular}[c]{@{}l@{}}Task: Determine the   text's sentiment polarity. You can select one of the following labels: \\      -1. negative, and 1. positive.\\      Text: {raw text}\end{tabular}                                                                                                                                                                                                                                                                                                                                                                                                                                                                                                                                                                                                                                                                                                                                                                                                                                                                                                                                                                                                                                                                                                                                                                                                                                                                                                                                                                                                                                                                                                                                                                                                                                                                                                                                                                                                                                                                                                                                                                                                                                                                                                                                                                                                                                                                                                                                              \\ \hline
XED-EC (multi)   \cite{ohman2020xed}            & \begin{tabular}[c]{@{}l@{}}Task: Determine the   emotion of the text based on eight discrete categories: 1. joy, 2. sadness,   3. anger, 4. fear, 5. surprise, \\6. disgust, 7. anticipation, 8. trust. \\      You have two options:\\      (a) Select one or more of the 8 emotions that represent the emotional state   depicted in the text. \\      (b) Choose ‘0. neutral’ if the text does not express any emotion.\\      Please provide your selection in the following format: {number}.   {emotion}.\\      Text: {raw text}\end{tabular}                                                                                                                                                                                                                                                                                                                                                                                                                                                                                                                                                                                                                                                                                                                                                                                                                                                                                                                                                                                                                                                                                                                                                                                                                                                                                                                                                                                                                                                                                                                                                                                                                                                                                                                                                                                                                                                                                                                                                                                               \\ \hline
MMS-SP (3)   \cite{augustyniak2023massively}      & \begin{tabular}[c]{@{}l@{}}Task: Determine the   text's sentiment polarity. You can select one of the following labels: \\      -1. negative, 0. neutral, and 1. positive.\\      Text: {raw text}\end{tabular}                                                                                                                                                                                                                                                                                                                                                                                                                                                                                                                                                                                                                                                                                                                                                                                                                                                                                                                                                                                                                                                                                                                                                                                                                                                                                                                                                                                                                                                                                                                                                                                                                                                                                                                                                                                                                                                                                                                                                                                                                                                                                                                                                                                                                                                                                                                                  \\ \hline
EMOTIC-EC (multi)   \cite{kosti2020context}       & \begin{tabular}[c]{@{}l@{}}<image>Task:   Determine the emotion(s) of the image based on a set of 26 discrete   categories using the image information. \\The list of emotions includes: \\      1. Happiness: feeling delighted; feeling enjoyment or amusement.\\      2. Sadness: feeling unhappy, sorrow, disappointed, or discouraged.\\      3. Anger: intense displeasure or rage; furious; resentful.\\      4. Fear: feeling suspicious or afraid of danger, threat, evil or pain;   horror.\\      5. Surprise: sudden discovery of something unexpected.\\      6. Aversion: feeling disgust, dislike, repulsion; feeling hate.\\      7. Excitement: feeling enthusiasm; stimulated; energetic.\\      8. Peace: well being and relaxed; no worry; having positive thoughts or   sensations; satisfied.\\      9. Affection: fond feelings; love; tenderness.\\      10. Annoyance: bothered by something or someone; irritated; impatient;   frustrated.\\      11. Anticipation: state of looking forward; hoping on or getting prepared   for possible future events.\\      12. Confidence: feeling of being certain; conviction that an outcome will   be favorable; encouraged; proud.\\      13. Disapproval: feeling that something is wrong or reprehensible;   contempt; hostile\\      14. Disconnection: feeling not interested in the main event of the   surrounding; indifferent; bored; distracted.\\      15. Disquietment: nervous; worried; upset; anxious; tense; pressured;   alarmed.\\      16. Doubt/Confusion: difficulty to understand or decide; thinking about   different options.\\      17. Embarrassment: feeling ashamed or guilty\\      18. Engagement: paying attention to something; absorbed into something;   curious; interested.\\      19. Esteem: feelings of favourable opinion or judgement; respect;   admiration; gratefulness.\\      20. Fatigue: weariness; tiredness; sleepy.\\      21. Pain: physical suffering.\\      22. Pleasure: feeling of delight in the senses.\\      23. Sensitivity: feeling of being physically or emotionally wounded;   feeling delicate or vulnerable.\\      24. Suffering: psychological or emotional pain; distressed;   anguished.\\      25. Sympathy: state of sharing others emotions, goals or troubles;   supportive; compassionate.\\      26. Yearning: strong desire to have something; jealous; envious; lust.\\      You can select one or more emotions from the list to represent the image's   emotional expression.\end{tabular} \\

\hline
\end{tabular}
\end{table*}

\begin{table*}[]
\footnotesize
\begin{tabular}{lp{15cm}}
\hline
{\color[HTML]{212121} Dataset}                     & Prompt                                                                                                                                                                                                                                                                                                                                                                                                                                                                                                                                                                                                                                                                                                                                                                                                                                                                                                                                                                                                                                                                                                                                                                                                                                                                                                                                                                                                                                                                                                                                                                                                                                                                                                                                                                                                                                                                                                                                                                                                                                                                                                                                                                                                                                                                                                                                                                                                                                                                                                                                           \\ \hline

EMOTIC-SI   \cite{kosti2020context}            & \begin{tabular}[c]{@{}l@{}}<image>Task:   Determine the valence, arousal, and dominance of the image using the image   information. The descriptions \\of these dimensions are as follows:\\      Valence: Measures how positive or pleasant an emotion is, ranging from   negative to positive.\\      Arousal: Measures the agitation level of the person, ranging from   calm/non-active to agitated/ready to act.\\      Dominance: Measures the level of control a person feels in the situation,   ranging from submissive/non-control to dominant/\\in-control.\\      Assign a real-valued score between -1 and 1 for each category:\\      -1: Indicates the most negative, calm/non-active, or   submissive/non-control.\\      1.0: Represents the most positive, agitated/ready to act, or   dominant/in-control.\\      Use the following template to report the intensity of each category:\\      valence: {intensity}, arousal: {intensity}, dominance: {intensity}.\end{tabular}                                                                                                                                                                                                                                                                                                                                                                                                                                                                                                                                                                                                                                                                                                                                                                                                                                                                                                                                                                                                                                                                                                                                                                                                                                                                                                                                                                                                                                                                                                                                                    \\ \hline
FER2013-EC (single)                         & \begin{tabular}[c]{@{}l@{}}<image>Task:   Determine the image's emotion using Ekman's categories: 1. happiness, 2.   sadness, 3. anger, 4. fear, 5. surprise, \\and 6. disgust.\\      You have two options: \\      (a) Select one Ekman label that best represents the emotional state depicted   in the image.\\      (b) Choose ‘0. neutral’ if the image does not express any emotion.\\      Please provide your selection in the following format: {number}. {emotion}.\end{tabular}                                                                                                                                                                                                                                                                                                                                                                                                                                                                                                                                                                                                                                                                                                                                                                                                                                                                                                                                                                                                                                                                                                                                                                                                                                                                                                                                                                                                                                                                                                                                                                                                                                                                                                                                                                                                                                                                                                                                                                                                                                                        \\ \hline
CFAPS-EC (single)   \cite{gong2011revision}      & \begin{tabular}[c]{@{}l@{}}<image>Task:   Determine the image's emotion using Ekman's categories: 1. happiness, 2.   sadness, 3. anger, 4. fear, 5. surprise, \\and 6. disgust.\\      You have two options: \\      (a) Select one Ekman label that best represents the emotional state depicted   in the image.\\      (b) Choose ‘0. neutral’ if the image does not express any emotion.\\      Please provide your selection in the following format: {number}. {emotion}.\end{tabular}                                                                                                                                                                                                                                                                                                                                                                                                                                                                                                                                                                                                                                                                                                                                                                                                                                                                                                                                                                                                                                                                                                                                                                                                                                                                                                                                                                                                                                                                                                                                                                                                                                                                                                                                                                                                                                                                                                                                                                                                                                                        \\ \hline
CFAPS-EI   \cite{gong2011revision}              & <image>Task:   Assess the intensity of emotion E in the image using a real number between 0   and 1, where 0 represents the least intensity and 1 represents the greatest   intensity. Emotion E: surprise.                                                                                                                                                                                                                                                                                                                                                                                                                                                                                                                                                                                                                                                                                                                                                                                                                                                                                                                                                                                                                                                                                                                                                                                                                                                                                                                                                                                                                                                                                                                                                                                                                                                                                                                                                                                                                                                                                                                                                                                                                                                                                                                                                                                                                                                                                                                                      \\ \hline
SANSEMO-EC (multi)   \cite{bujnowski2024samsemo}   & \begin{tabular}[c]{@{}l@{}}<video>Task:   Determine the video's emotion based on its content (text transcribed from the   video's audio) and video \\information, using Ekman's categories: 1. happiness,   2. sadness, 3. anger, 4. fear, 5. surprise, and 6. disgust. \\      You have three options: \\      (a) Select up to two Ekman labels for a given scene.\\      (b) Choose ‘0. neutral’ if the video does not express any emotion.\\      (c) Select '99. other emotions' if the basic six Ekman emotions are   inadequate. \\      Please provide your selection in the following format: {number}.   {emotion}.\\      Content: {transcription}\end{tabular}                                                                                                                                                                                                                                                                                                                                                                                                                                                                                                                                                                                                                                                                                                                                                                                                                                                                                                                                                                                                                                                                                                                                                                                                                                                                                                                                                                                                                                                                                                                                                                                                                                                                                                                                                                                                                                                                         \\ \hline
CHSIMS-SP (3)   \cite{yu2020ch}                  & \begin{tabular}[c]{@{}l@{}}<video>Task:   Determine the video's sentiment polarity according to its content (the text   transcribed from the video's audio) \\and video information. You can select one   of the following labels: \\      -1. negative, 0. neutral, and 1. positive.\\      Content: {transcription}\end{tabular}                                                                                                                                                                                                                                                                                                                                                                                                                                                                                                                                                                                                                                                                                                                                                                                                                                                                                                                                                                                                                                                                                                                                                                                                                                                                                                                                                                                                                                                                                                                                                                                                                                                                                                                                                                                                                                                                                                                                                                                                                                                                                                                                                                                                                 \\ \hline
CHSIMS-SI   \cite{yu2020ch}                   & \begin{tabular}[c]{@{}l@{}}<video>Task:   Determine the video's sentiment strength according to its content (the text   transcribed from the video's audio) \\and video information. Choose one of the   following values: \\      {-1.0, -0.8, -0.6, -0.4, -0.2, 0.0, 0.2, 0.4, 0.6, 0.8, 1.0}, where -1.0   indicates the most negative sentiment, 0.0 represents neutral, and \\1.0   indicates the most positive sentiment.\\      Content: {transcription}\end{tabular}                                                                                                                                                                                                                                                                                                                                                                                                                                                                                                                                                                                                                                                                                                                                                                                                                                                                                                                                                                                                                                                                                                                                                                                                                                                                                                                                                                                                                                                                                                                                                                                                                                                                                                                                                                                                                                                                                                                                                                                                                                                                         \\ \hline
MELD-EC (single)   \cite{poria2019meld}         & \begin{tabular}[c]{@{}l@{}}<video>Task:   Determine the video's emotion based on its content (text transcribed from the   video's audio) and video information\\, using Ekman's categories: 1. happiness,   2. sadness, 3. anger, 4. fear, 5. surprise, and 6. disgust. \\      You have two options: \\      (a) Select one Ekman label that best represents the emotional state   depicted in the video.\\      (b) Choose ‘0. neutral’ if the video does not express any emotion.\\      Please provide your selection in the following format: {number}.   {emotion}.\\      Content: {transcription}\end{tabular}                                                                                                                                                                                                                                                                                                                                                                                                                                                                                                                                                                                                                                                                                                                                                                                                                                                                                                                                                                                                                                                                                                                                                                                                                                                                                                                                                                                                                                                                                                                                                                                                                                                                                                                                                                                                                                                                                                                             \\ \hline
MELD-SP (3)   \cite{poria2019meld}                   & \begin{tabular}[c]{@{}l@{}}<video>Task:   Determine the video's sentiment polarity according to its content (the text   transcribed from the video's audio) \\and video information. You can select one   of the following labels: \\      -1. negative, 0. neutral, and 1. positive.\\      Content: {transcription}\end{tabular}                                                                                                                                                                                                                                                                                                                                                                                                                                                                                                                                                                                                                                                                                                                                                                                                                                                                                                                                                                                                                                                                                                                                                                                                                                                                                                                                                                                                                                                                                                                                                                                                                                                                                                                                                                                                                                                                                                                                                                                                                                                                                                                                                                                                                 \\ \hline
\end{tabular}
\end{table*}

\end{document}